\documentclass{CUP-JNL-DTM}%
\usepackage{graphicx}
\usepackage{multicol,multirow}
\usepackage{amsmath,amssymb,amsfonts}
\usepackage{mathrsfs}
\usepackage{amsthm}
\usepackage{rotating}
\usepackage{appendix}
\usepackage{nomencl}
\usepackage{etoolbox}
\renewcommand\nomgroup[1]{%
  \item[\bfseries
  \ifstrequal{#1}{S}{Symbols}{%
  \ifstrequal{#1}{M}{Matrix/tensors}{%
  \ifstrequal{#1}{N}{Non-dimensional group}{%
  \ifstrequal{#1}{A}{Acronyms}{}}}}%
]}

\usepackage[numbers]{natbib}
\usepackage{ifpdf}
\usepackage[T1]{fontenc}
\usepackage{newtxtext}
\usepackage{newtxmath}
\usepackage{textcomp}
\usepackage{xcolor}
\usepackage{lipsum}
\usepackage[colorlinks,allcolors=blue]{hyperref}
\usepackage{array}

\usepackage{tikz}
\usetikzlibrary{shapes}
\usepackage{soul}
\usepackage{subcaption}
\usepackage{hyperref}
\hypersetup{
    colorlinks = true,
    urlcolor   = blue,
    citecolor  = black,
}
\usepackage{placeins}
\usepackage{mleftright}

\theoremstyle{definition}

\numberwithin{equation}{section}

\def\high#1{\textsuperscript{#1}}

\def\vec#1{\boldsymbol{#1}}
\def\mat#1{\pmb{#1}}
\def\sym#1{\mbox{\textit{#1}}}

\usepackage{color}

\usepackage{listings}
\usepackage{algorithm2e}

\articletype{RESEARCH ARTICLE}
\jyear{YEAR}
\begin{document}
\begin{Frontmatter}

\title[Article Title]{Decoder decomposition for the analysis of the latent space of nonlinear autoencoders with wind-tunnel  experimental data}

\author[1]{Yaxin Mo}
\author[2,1]{Tullio Traverso}
\author[1,2,3]{Luca Magri}
\authormark{Yaxin Mo \textit{et al}.}
\address[1]{
    \orgdiv{Department of Aeronautics}, 
    \orgname{Imperial College London}, 
    \orgaddress{\city{London}, \postcode{SW7 2AZ},  \country{UK}}
    \email{l.magri@imperial.ac.uk}
}

\address[2]{
    \orgname{The Alan Turing Institute}, 
    \orgaddress{\city{London}, \postcode{NW1 2DB},  \country{UK}}
}
\address[3]{
    \orgname{Politecnico di Torino, DIMEAS}, 
    \orgaddress{\city{Torino}, \postcode{10129},  \country{Italy}}
}


\abstract{
Turbulent flows are chaotic and multi-scale dynamical systems, which have large numbers of degrees of freedom.  
Turbulent flows, however, can be modelled with a smaller number of degrees of freedom when using the appropriate coordinate system, which is the goal of dimensionality reduction via nonlinear autoencoders. Autoencoders are expressive tools, but they are difficult to interpret. 
The goal of this paper is to propose a method to aid the interpretability of autoencoders. This is the {\it decoder decomposition}.
First, we propose the decoder decomposition, which is a post-processing method to connect the latent variables to the coherent structures of flows. 
Second, we apply the decoder decomposition to analyse the latent space of synthetic data of a two-dimensional unsteady wake past a cylinder. 
We find that the dimension of latent space has a significant impact on the interpretability of autoencoders. We identify the physical and spurious latent variables. 
Third, we apply the decoder decomposition to the latent space of wind-tunnel experimental data of a three-dimensional turbulent wake past a bluff body. 
We show that the reconstruction error is a function of both the latent space dimension and the decoder size, which are correlated. 
Finally, we apply the decoder decomposition to rank and select latent variables based on the coherent structures that they represent. This is useful to filter unwanted or spurious latent variables, or to pinpoint specific the coherent structures of interest.
The ability to rank and select latent variables will help users design and interpret nonlinear autoencoders. 
}

\end{Frontmatter}
\section*{Impact Statement}
Nonlinear dimensionality reduction by autoencoders can efficiently compress high-dimensional data into a low-dimensional latent space, but the results may be difficult to interpret.
We propose the {\it decoder decomposition} to select and rank the latent variables based on the coherent structures that they represent. This opens opportunities for building interpretable models with nonlinear autoencoding.

\section{Introduction}\label{sec:intro}
Turbulent flows are nonlinear and multi-scale systems, which have large numbers of degrees of freedom. 
High-fidelity simulations of turbulent flows can be performed by solving the governing equations on fine spatiotemporal grids, but the computational cost can be prohibitively high \citep[e.g.,][]{rowley2017ModelReductionFlow}.

When computationally cheaper modelling of turbulent flows is needed, reduced-order models are applied to approximate the flows with fewer  degrees of freedom \citep[e.g.,][]{noack2011rom_for_control,rowley2017ModelReductionFlow}.
Commonly, reduced-order models are constructed via projection-based methods, such as proper orthogonal decomposition (POD) \citep[e.g.,][]{noack2011rom_for_control,rowley2017ModelReductionFlow}.
POD is a decomposition method dating back to 1970 \citep{lumley1970StochasticToolsTurbulence}, which enables the computation of an optimal linear subspace  based on the energy norm \citep[e.g.,][]{taira2017modal}.
Each POD mode is associated with an energy, which ranks the importance of the mode, and 
POD modes can be interpreted as coherent structures of flows \citep[e.g.,][]{alfonsi2007structure-turbulent-boundary,kevlahan1994comparison_technique_identify_structure,rigas2014coherent_structure_wake}.
There are other methods of linear decomposition, such as the spectral POD \citep{lumley1970StochasticToolsTurbulence,schmidt2020guide_spod}, dynamic mode decomposition \citep{schmid2010DynamicModeDecomposition,RefWorks:RefID:84-h.2014dynamic}, and wavelet analysis \citep{RefWorks:RefID:89-albukrek2002divergence-free}, which have been employed for dimensionality reduction and discovery of coherent structures. 
These linear methods are relatively straightforward to implement, but may require large numbers of modes to reduce the approximation error for accurate modelling of nonlinear flows~\citep{alfonsi2007structure-turbulent-boundary,murata2020nonlinear}.
On the other hand, nonlinear reduced-order modelling  seeks nonlinear manifolds onto which perform dimensionality reduction to approximate the dynamics~\citep[e.g.,][]{magril.2022InterpretabilityProperLatent,racca2023predicting}. \\

Machine learning has been increasingly applied for dimensionality reduction of fluids, in particular via nonlinear autoencoders~\citep{csala2022ComparingDifferentNonlinear,doan2023ConvolutionalAutoencoderSpatiotemporal,eivazi2022extract_orthogonal_ae_modes,fukami2021AEandROM,magril.2022InterpretabilityProperLatent}.
Autoencoders consist of an encoding and a decoding part: the encoding part maps the input (the physical flow field) into a lower-dimensional latent space, whereas the decoding part maps the latent space back to the physical space. 
The purpose of the autoencoders is to approximate the identity mapping.
The turbulent system's dynamics can be predicted on the low-dimensional latent space with sequential methods such as reservoir computers \cite{RefWorks:RefID:6-doan2021auto-encoded, racca2023predicting} and Long short term memory networks \cite{RefWorks:RefID:39-nakamura2020cnn-ae/lstm} for flow forecasting. 
The latent space of autoencoder, which is a nonlinear manifold~\citep{magril.2022InterpretabilityProperLatent}, may be difficult to interpret. 
This is because the latent variables are entangled and their coordinate bases may not be locally orthogonal~\citep{magril.2022InterpretabilityProperLatent}. 

The mode-decomposing autoencoder (MD-AE) was developed by \citet{murata2020nonlinear} to improve the interpretability of nonlinear decomposed fields. 
MD-AEs assign a decoder to each variable of the latent space vector (latent variable), and then superpose the single-decoder outputs to generate the MD-AE output. 
MD-AEs improve the visualization of flow decomposition because the effect of each latent variable is isolated in the decoding part, unlike standard AEs in which one decoder contains the effect of all latent variables.
With  the same latent space dimension, however, an AE  outperforms an MD-AE in terms of reconstruction error \citep{eivazi2022extract_orthogonal_ae_modes,murata2020nonlinear}.
This is because the design of  MD-AEs does not capture the nonlinear coupling of the decoded latent variables. 
For highly nonlinear flows, the difference between the reconstruction errors may be significant \citep{csala2022ComparingDifferentNonlinear}.
Other autoencoder architectures, such as the hierarchical autoencoder \citep{fukami2020hierarchical_ae} and the $\beta$-variational autoencoder \citep{eivazi2022extract_orthogonal_ae_modes}, have been proposed to improve the interpretability of the decomposed fields. The decomposed fields, nonetheless, may be still influenced by more than one latent variable, so the exact effect of each latent variable on the output is difficult to isolate.
A comparison between mode decomposition methods, including MD-AE and POD, can be found in the work by \citet{csala2022ComparingDifferentNonlinear}.
The interpretability of autoencoder latent space is an open problem \citep{magril.2022InterpretabilityProperLatent,vinuesa2021interpretableAI}.\\ 

The overarching objective of this paper is to aid the interpretability of the latent space of two common autoencoders. 
For this purpose, we propose the {\it decoder decomposition}, which is a post-processing method for raking and selecting the latent space of nonlinear autoencoders. 
The decoder decomposition is based on POD modes and the decoders' sensitivities computed as  gradients. 
Specifically, the objectives are 
(i) propose the decoder decomposition, which disentangles the contribution of latent variables in the decoded field; 
(ii) analyse and verify the decoder decomposition of two commonly used autoencoders (the standard autoencoder and the mode-decomposing autoencoder) with synthetic data from numerical simulation of the unsteady two-dimensional wake past a cylinder; and 
(iii) apply the decoder decomposition to gain physical insight into a realistic flow (three-dimensional turbulent flow from by wind-tunnel experiments) and isolate the latent variables of physical interest with a filtering strategy. The three-dimensional turbulent flow is the high-speed wake past a bluff body. \\

The paper is organized as follows:
First, we provide an overview of the proper orthogonal decomposition (POD) in Section~\ref{sec:method-pod} and detail our datasets in Section~\ref{sec:data}.
Next, we introduce the autoencoder architectures in Section~\ref{sec:all_ae} and propose the decoder decomposition in Section~\ref{sec:method-decoderdecomposition}.
We apply the decoder decomposition to decompose the unsteady laminar cylinder wake and investigate the impact of the dimension of the latent space on the interpretability of autoencoders (AEs) in Section~\ref{sec:results-cylinder}.
We decompose the wind-tunnel wake in Section~\ref{sec:results-exp} and demonstrate how to rank and select latent variables to filter for coherent structures in the output of the AEs.
We present our conclusions in Section~\ref{sec:conclusion}.

\section{Proper Orthogonal Decomposition}\label{sec:method-pod}
Let $\mat{Q} \in \mathbb{R}^{N\times N_t}$ be a generic dataset of a measured fluctuating quantity from a flow field such that each snapshot is a column, where $N$ is the product of the number of grid points, the number of variables, and $N_t$ is the number of snapshots.
The $i$-th row, $\vec{Q}_{i,:}$, is the time series of the measured quantity at grid point $i$.
The $t$-th column, $\vec{Q}_{:,t}$, is the snapshot of the measured quantities at the discrete time step $t$.
(In this paper, either the fluctuating velocities or the pressure is measured.)
The covariance matrix $\pmb{C}\in \mathbb{R}^{N\times N}$ is
\begin{equation}
    \mat{C} = 
    \frac{1}{N_t-1} \mat{W}_p\mat{Q}\mat{Q}^T\mat{W}_p^T = 
    \mat{\Phi\Lambda\Phi}^T, 
    \label{eq:method-POD-C}
\end{equation}
where $\mat{W}_p$ is the POD weight matrix given to each element of $\mat{Q}$, $\mat{\Lambda}$ is the diagonal matrix of the eigenvalues and $\mat{\Phi} \in \mathbb{R}^{N\times N}$ is the matrix of the eigenvectors.
The weight matrix for each dataset is given in section~\ref{sec:data}.
The $i$-th column of the matrix of eigenvalues, $\vec{\Phi}_{:,i}$, is the POD mode~$i$, which represents the $i$-th principal axis of the space occupied by the observed data ranked by the eigenvalues, i.e., the energy of the mode \citep{taira2017modal,weiss2019tutorialpod}. 
The matrix of temporal coefficients $\mat{A}$ 
is obtained by projecting the data matrix onto the POD modes
\begin{equation}
    \mat{A} = \mat{Q}^T \mat{\Phi} , 
    \label{eq:method-POD-A}
\end{equation}
which are the temporal coordinates of the data snapshots along the principal axes \citep{weiss2019tutorialpod}. 
The $i$-th row of $\mat{A}$, denoted $\vec{A}_{i,:}$, contains the time series of the time coefficient of the $i$-th POD mode. 
The time coefficient for the mode $i$ at time step $t$ is $A_{i,t}$.
By setting $N_m < N$ as the truncated number of modes for reconstructing the flow, we approximate the flow field in the subspace spanned by the first $N_m$ POD modes
\begin{equation}
    \tilde{\mat{Q}} = \sum^{N_m}_{i=1}  \vec{\Phi}_{:,i} \vec{A}_{i,:}^T.
    \label{eq:method-pod_decomposed_fields}
\end{equation}
If $N_m=N$, then $\tilde{\mat{Q}}=\mat{Q}$, i.e., no approximation is made, and only a linear change of coordinate system is performed.

\section{Datasets and pre-processing}\label{sec:data}
Two datasets are considered in this paper.
First, an unsteady laminar wake behind a cylinder, which is the benchmark case whose dynamics are well-known (section~\ref{sec:data-cylinder}).
Second, an wind-tunnel dataset of a three-dimensional turbulent bluff body wake~\cite{rowand.brackston2017FeedbackControlThreeDimensional} (section~\ref{sec:data-exp}).

\subsection{Unsteady laminar wake of a two-dimensional cylinder}\label{sec:data-cylinder}
\begin{figure}
    \centering
    \FIG{
        \includegraphics[width=0.8\textwidth]{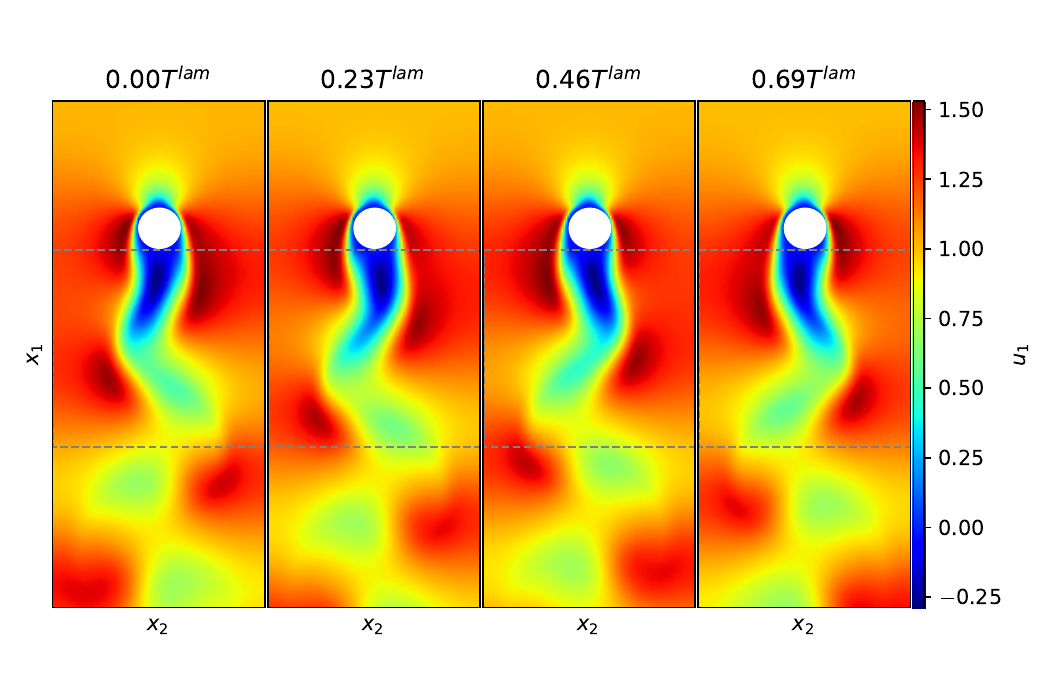}
    }{
        \caption{Snapshots of the streamwise velocity of the laminar wake dataset at different times within the same period. A vortex shedding period is denoted with $T^{\sym{lam}}$. The area bounded by the grey box is used for training.} 
    }\label{fig:cylinder-domain}
\end{figure}

The unsteady laminar wake behind a 2D circular cylinder at Reynolds number $\sym{Re}=100$ is generated by solving the dimensionless Navier-Stokes Equations
\begin{equation}
    \begin{cases}
        \nabla \cdot \vec{u} = 0 \\
        \frac{d\vec{u}}{d t} + \vec{u} \cdot \nabla \vec{u} = -\nabla p + \frac{1}{\sym{Re}}\triangle\vec{u}
    \end{cases},
\end{equation}
where the vector $\vec{u} \in \mathbb{R}^{N_u}$ with $N_u=2$ is the velocity, $p$ is the pressure, and $t$ is the time.
The velocity and length are non-dimensionalised by the inlet velocity $U_\infty$ and the diameter of the cylinder $D$. 
The computation domain has size $L_1=12$, $L_2=5$ and $L_3=1$ (Figure~\ref{fig:cylinder-domain}), divided uniformly into $513$, $129$ and $8$ nodes in streamwise, wall-normal and spanwise directions, respectively. 
The boundary conditions are  
Dirichlet boundary condition at $x_1=12$;
slip walls at $x_2=0$ and $x_2=5$;
and the periodic at $x_3=0$ and $x_3=1$.
The centre of the cylinder is at $(3,2.5)$.
The dataset is simulated with direct numerical simulation using Xcompact3D \citep{bartholomew2022Xcompact3dIncompact3d}.
A time step $\Delta t = 0.0002$ is chosen to satisfy the Courant–Friedrichs–Lewy condition.
The numerical schemes are the 6\high{th}-order compact scheme in space \citep{laizet2009HighorderCompactSchemesa} and the 3\high{rd}-order Adams-Bashforth in time.
The simulation matches the results of \citet{gangaprasath2014EffectsAspectRatio}.

The transient period of the first $100$ time units at the beginning of the simulation is discarded to ensure the dataset contains only the periodic flow \citep{gangaprasath2014EffectsAspectRatio}.
Snapshots of the streamwise velocity, $u_1$, are saved every $0.125$ time units, giving over 40 snapshots per vortex shedding period, $T^{\sym{lam}}$
The final dataset contains 720 snapshots. 
We consider a 200-by-129 grid behind the cylinder (grey box in Figure~\ref{fig:cylinder-domain}).
The domain includes areas from $0$ to $4.5D$ downstream of the body, capturing the near wake and the vortex shedding \citep{m.m.zdravkovich1997FlowCircularCylinders}.
The dataset used by the AEs, $\mat{U} \in \mathbb{R}^{N_1 \times N_2 \times N_t}$, consists of the fluctuating streamwise velocity, where $N_t=720$ is the number of snapshots, $N_1=200$ and $N_1=129$ are the number of grid points in the streamwise and wall-normal directions, respectively.

\begin{figure}
    \centering
    \includegraphics[width=\textwidth]{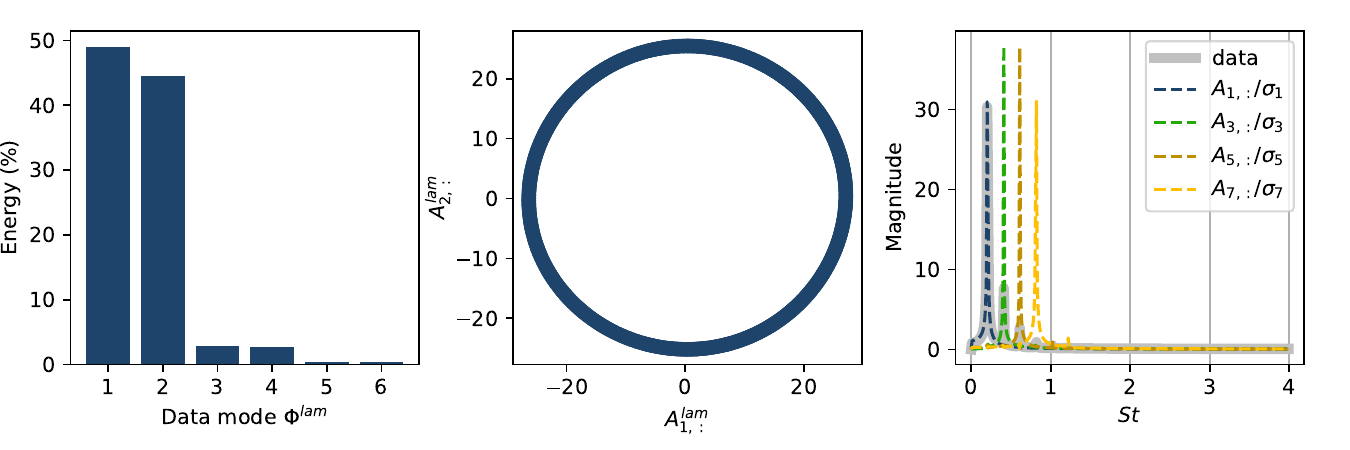}
    \caption{
        POD of the laminar wake $\mat{U}$.
        Left: the percentage energy contained in the first six POD modes of the unsteady wake $\mat{\Phi}^{\sym{lam}}$ (data mode). Data modes 1\&2, 3\&4 and 5\&6 contain similar flow energy and oscillate at the same frequency but out of phase.
        Centre: phase plot of the first two data time coefficients. 
        Right: the frequency spectrum of the data and the data time coefficients 1, 3, 5 and 7, normalized by the standard deviation. The data contains the vortex shedding frequency and its harmonics. (Since each pair has the same frequency spectrum, only the odd data modes are shown here).
    }\label{fig:cylinder-data-pod-results}
\end{figure}

We perform POD on the dataset $\mat{U}$ to obtain the POD modes $\mat{\Phi}^{\sym{lam}}$ and the matrix of time coefficients $\mat{A}^{\sym{lam}}$, referred to as the ``data modes'' and ``data time coefficients'', respectively, where $\sym{lam}$ stands for ``laminar''.
Examination of the eigenvalues shows that the data modes are out-of-phase pairs, the two modes in a pair contain a similar amount of flow energy (Figure~\ref{fig:cylinder-data-pod-results}, left panel) and their time coefficients are $90^\circ$ out-of-phase (Figure~\ref{fig:cylinder-data-pod-results}, centre panel).
The first two data modes represent the vortex shedding, at the vortex shedding frequency, and the higher modes oscillate at the harmonic frequencies \citep{loiseau2020pod-galerkin}.
The magnitude of the fast Fourier transform in time, summed over the grid points, is used to measure the overall frequency content of the dataset, shown in Figure~\ref{fig:cylinder-data-pod-results} (right panel).
The frequency content of the flow is described in terms of the Strouhal number, $\sym{St} = f D/U_\infty$, where $f$ is the frequency measured in Hz.
The dominant frequency is the vortex shedding frequency at $St=0.23$, and the dataset also contains higher harmonic frequencies. 
Each pair of data modes oscillates at a single, distinct frequency (Figure~\ref{fig:cylinder-data-pod-results}, right panel).
The first pair, comprised of data modes 1 and 2, oscillate at the vortex shedding frequency, while the subsequent pairs, composed of data modes 3 and 4, oscillate at the first harmonic, and so forth.
The concentration of the flow energy in the leading data modes and the distinct frequencies of the data time coefficients make the unsteady wake behind a cylinder suitable for dimensionality reduction and interpretation with POD.\@

\subsection{Turbulent wake of a bluff body from wind tunnel experiments}\label{sec:data-exp}

\begin{figure}
    \centering
    \includegraphics[width=\textwidth]{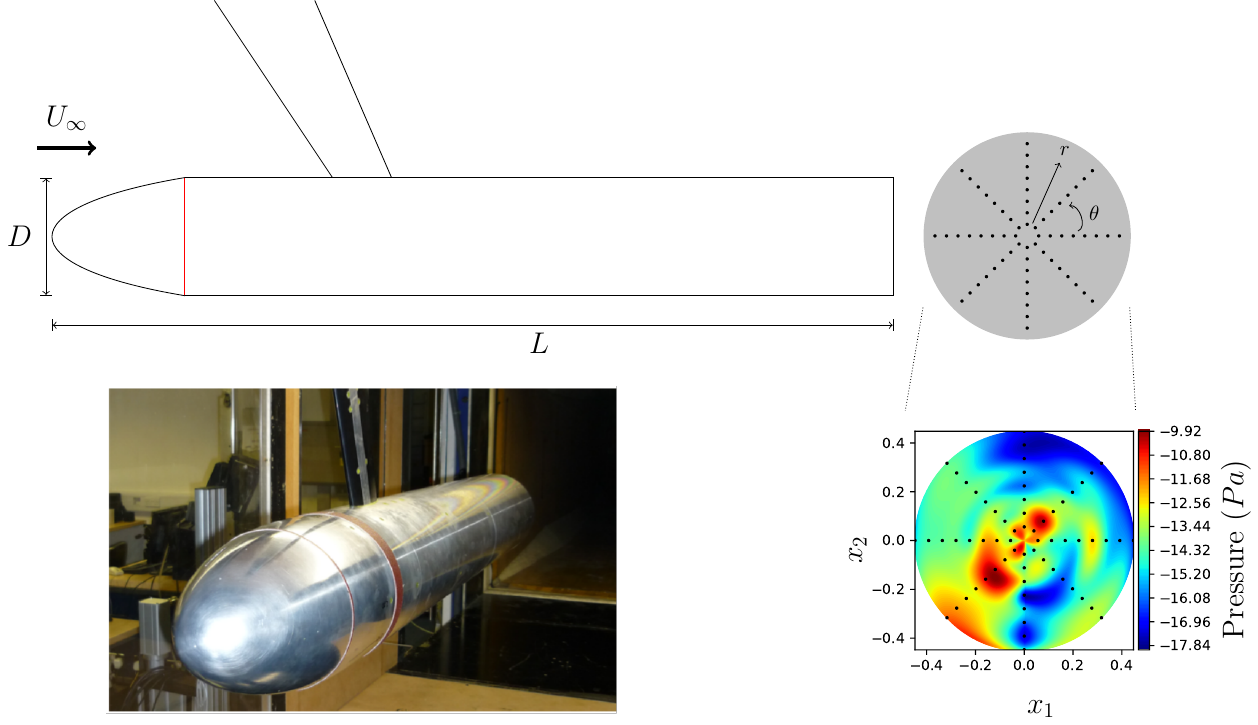}
    \caption{
        Experimental set-up, reproduced from \citep{rigas2021experiment}.
        The dimensions $x_1$ and $x_2$ are the measured location nondimensionalized by the diameter $D$.
        The black dots mark the location of the pressure sensors. 
    }\label{fig:exp-data-rig}
\end{figure}
The turbulent wake of an axisymmetric body from wind tunnel experiments \citep{rigas2014coherent_structure_wake,rowand.brackston2017FeedbackControlThreeDimensional} is employed in the second part of this paper.
Figure~\ref{fig:exp-data-rig} shows the experimental setup. 
The axisymmetric body has a diameter of $D=196.5$~mm and a length-to-diameter ratio $L/D = 6.5$.
Pressure measurements at the base of the body are collected via 64 static pressure sensors placed on a regular polar grid, with eight grid points in the radial direction and eight grid points in the azimuthal direction (Figure~\ref{fig:exp-data-rig}).
We provide here an overview of the dataset.
For a more in-depth analysis of the dataset and the experimental setup, the readers are referred to \citet{rigas2014coherent_structure_wake}.

\begin{figure}
    \centering
    \includegraphics[width=\textwidth]{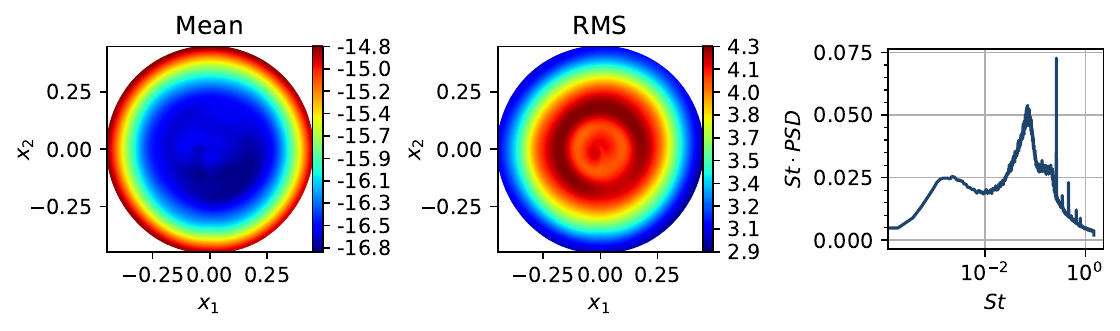}
    \caption{
        The wind-tunnel pressure dataset $\mat{P}$.
        Left: Mean pressure.
        Centre: RMS pressure.
        Right: The premultiplied PSD ($\sym{St} \cdot$ PSD) of the wind-tunnel dataset, with peaks at $\sym{St} \approx 0.002, 0.06$ and $\sym{St} \approx 0.2$ and its harmonics. The peaks correspond to the three-dimensional rotation of the wake, the pulsation of the vortex core, and the vortex shedding and its harmonics, respectively.
    }\label{fig:exp-data-overall}
\end{figure}

The mean and root-mean-squared (RMS) pressures are both axisymmetric (Figure~\ref{fig:exp-data-overall}).
The power spectral density (PSD) of the pressure data is shown in Figure~\ref{fig:exp-data-overall} (right), premultiplied by the Strouhal number to improve the visualization~\citep{rigas2014coherent_structure_wake}.
The peaks at $\sym{St}\approx 0.002$, $0.06$ and $0.2$ are associated with the 3D rotation of the symmetry plane, pulsation of the vortex core and vortex shedding, respectively.

\begin{figure}
    \centering
    \includegraphics[width=0.8\textwidth]{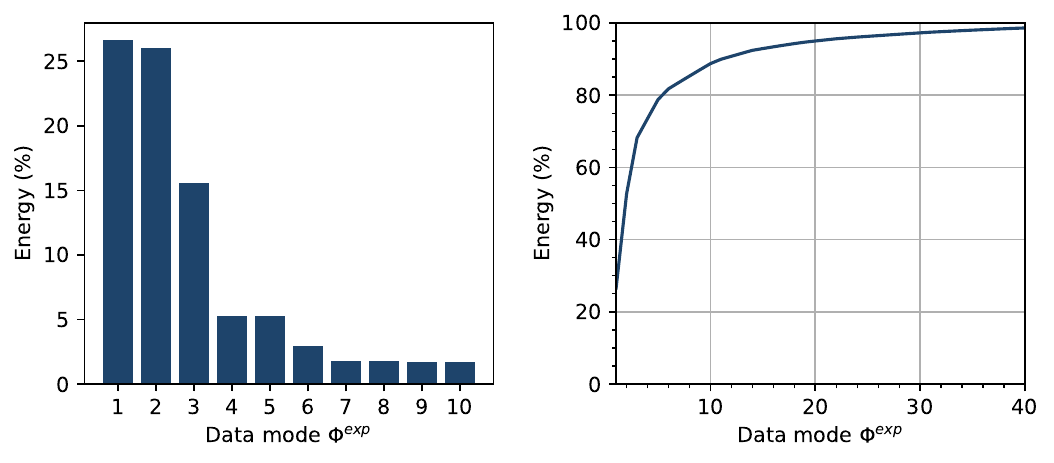}
    \caption{
        POD of the wind-tunnel dataset $\mat{P}$. 
        Left: percentage energy of the first 10 data modes.
        Right: cumulative percentage energy of POD modes. The reconstruction of the pressure dataset to 95\% energy needs 21 data modes.
    }\label{fig:exp-data-lam}
\end{figure}

\begin{figure}
    \centering
    \includegraphics[width=0.6\textwidth]{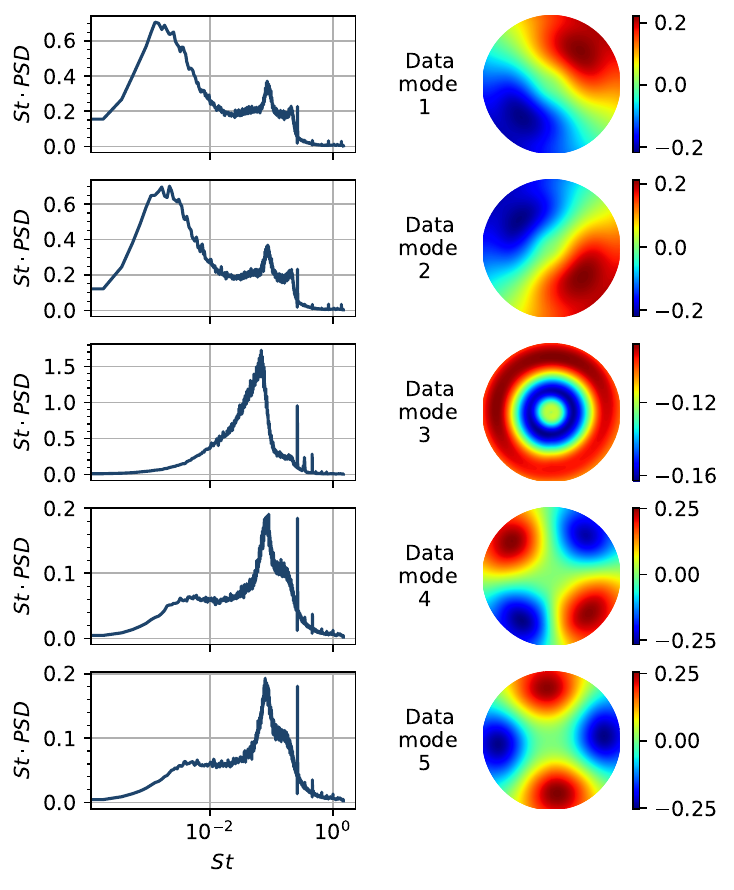}
    \caption{
    Right: The first five data modes of the wind-tunnel dataset $\vec{\Phi}^{\sym{exp}}_{:,1}$ to $\vec{\Phi}^{\sym{exp}}_{:,5}$. 
    Left: the premultiplied PSD ($St \cdot$ PSD) of their associated time coefficients $\vec{A}^{\sym{exp}}_{1,:}$ to $\vec{A}^{\sym{exp}}_{5,:}$. 
    }\label{fig:exp-data-pod-modes}
\end{figure}

The weighted POD (each data point is weighted by the area of the element of the polar grid) is applied to the dataset \citep{rigas2014coherent_structure_wake, rowand.brackston2017FeedbackControlThreeDimensional}.
The resulting POD modes $\mat{\Phi}^{\sym{exp}}$ and time coefficients $\mat{A}^{\sym{exp}}$ are the data modes and data time coefficients, respectively, where $\sym{exp}$ stands for ``experimental''.
Figure~\ref{fig:exp-data-lam} shows the energy contained in each data mode of the wind-tunnel dataset and the cumulative energy of the modes.
The flow energy is spread over more modes than the laminar case, and 21 data modes are needed to reconstruct the dataset to recover 95\% of the energy.
The modes and the premultiplied PSD of their time coefficients are shown in Figure~\ref{fig:exp-data-pod-modes}.
Data modes~1 and~2 are antisymmetric and have frequency peaks at $St \approx 0.002, 0.1 \text{ and } 0.2$.
As a pair, data modes 1 and 2 represent vortex shedding and the `very-low-frequency' 3D rotation of the symmetry plane \citep{rigas2014coherent_structure_wake}.
Data mode 3 has a frequency peak at $St \approx 0.06$ and is associated with the pulsation of the vortex core \citep{rigas2014coherent_structure_wake,rowand.brackston2017FeedbackControlThreeDimensional}.
The data modes 4 and 5 contain the subharmonic frequencies.
Data modes 1 to 5 contain the coherent structures of the dataset.
We base our analysis and latent variable selection on the interpretation of the data modes $\mat{\Phi}^{\sym{exp}}$ provided here.
We consider the fluctuating field, following the same logic as for the laminar cylinder wake dataset.
The final experimental pressure dataset employed for training is $\mat{P} \in \mathbb{R}^{N \times N_t}$, where $N_t=420000$ is the number of available snapshots in time, sampled at a sampling frequency $225$ Hz and $N=64$ is the number of pressure sensors. 
The dataset $\mat{P}$ contains data collected for over 1800~s.

\section{Autoencoder architectures}\label{sec:all_ae}
An autoencoder approximates the identity function by means of an encoder and a decoder. 
We define a generic $\mat{Y} \in \mathbb{R}^{N^* \times N_t}$, where $N^*$ is the spatial dimension.
For all time steps, the encoder $F_{en}: \mathbb{R}^{N^*} \rightarrow \mathbb{R}^{N_z}$ maps the snapshot of the data at time $t$, $\vec{Y}_{:,t}$ into a latent space, represented by a latent vector $\vec{Z}_{:,t}$, where $N_z \ll N^*$ is the dimension of the latent space.
The $i$-th latent variable is $\vec{Z}_{i,:}$.
The encoder, $F_{en}$, is a composition of layers $f^{(1)}_{en}, f^{(2)}_{en}, \dots, f^{(n)}_{en}$ and  activation functions $\theta^{(1)}_{en}, \theta^{(2)}_{en}, \dots, \theta^{(n)}_{en}$ applied to each layer.
The decoder, $F_{de}$, is a composition of  layers $f^{(1)}_{de}, f^{(2)}_{de}, \dots, f^{(n)}_{de}$ and the activation functions $\theta^{(1)}_{de}, \theta^{(2)}_{de}, \dots, \theta^{(n)}_{de}$ applied to each layer 
\begin{equation}
\begin{aligned}
    F_{en} &= \theta^{(n)}_{en} \circ f^{(n)}_{en} \circ \theta^{(n-1)}_{en} \circ f^{(n-1)}_{en} \dots \circ \theta^{(1)}_{en} \circ f^{(1)}_{en}, \\
    F_{de} &= \theta^{(n)}_{de} \circ f^{(n)}_{de} \circ \theta^{(n-1)}_{de} \circ f^{(n-1)}_{de} \dots \circ \theta^{(1)}_{de} \circ f^{(1)}_{de}, 
\end{aligned}
\end{equation}
where $n$ is the number of layers.
Each layer ($f^{(*)}_{en}$ or $f^{(*)}_{de}$) maps the output of the previous layer to the input of the next layer, and is a function of the trainable parameters ($\omega^{(*)}_{en}$ or $\omega^{(*)}_{de}$).
Each layer is either a convolution or a linear mapping depending on the test case and the datasets.
Details of layers and activation functions are explained in sections~\ref{sec:method-network-ae} and~\ref{sec:method-network-mdae}.

The decoding part maps the instant latent vector at $t$ back to the original space, with the output $\hat{\vec{Y}}_{:,t}$ approximating the input $\vec{Y}_{:,t}$.
The autoencoder, comprised of the encoding and decoding parts, is trained with all $t$ to output $\hat{\mat{Y}}$, which approximates the input $\mat{Y}$.
The error between $\hat{\mat{Y}}$ and $\mat{Y}$ is measured with the mean squared error (MSE) 
\begin{equation}
    \sym{MSE}\left( \mat{Y}, \hat{\mat{Y}} \right) = \frac{1}{N^* \times N_t} \sum^{N^*}_{i=1}\sum^{N_t}_{t=1} (Y_{i,t} - \hat{Y}_{i,t})^2.
    \label{eq:mse}
\end{equation}
The parameters of the autoencoder, $\vec{\omega} = \{ \omega^{(1)}_{en}, \dots, \omega^{(n)}_{en}, \omega^{(1)}_{de}, \dots, \omega^{(n)}_{de}, \}$,  are obtained by minimizing the MSE
\begin{equation}
    \vec{\omega}^* = \arg\min_{\boldsymbol{\omega}} \sym{MSE}\left( \mat{Y}, \hat{\mat{Y}} \right).
    \label{eq:trainable-parameters}
\end{equation}

This analysis includes two different autoencoder architectures --- the standard autoencoder (AE) (section~\ref{sec:method-network-ae}) and the mode-decomposing autoencoder (MD-AE) (section~\ref{sec:method-network-mdae}).
To compare the results with linear decomposition methods (POD), we use networks with no bias and train on the fluctuating fields, so our results are one-to-one comparable to those obtained with POD.
Two types of intermediate layers are employed --- convolutional layers for the laminar dataset, and feedforward layers for the wind-tunnel dataset to handle the polar grid.

In this section, we define the autoencoders for a generic input dataset $\mat{Y}$ where the spatial dimension $N^*$ depends on the dataset under investigation  (numerical and wind-tunnel datasets) and the autoencoder architectures.
The inputs and autoencoder architectures are summarized in Table~\ref{tab:test-input-dimension}.
\begin{table}
    \centering
    \caption{
        Different autoencoder architectures and the training datasets.
    }\label{tab:test-input-dimension}
    \begin{tabular}{|c|c|c|c|}
        \hline
        \textbf{Results in section} & \textbf{Autoencoder architecture} & \textbf{Inputs $\mat{Y}$} & \textbf{Spatial dimension $N^*$} \\
        \hline
        \ref{sec:result-cylinder-mdae} & MD-AE (CNN) & \mat{U} & $200 \times 129$ \\
        \ref{sec:result-cylinder-ae} & AE (CNN) & \mat{U} & $200 \times 129$ \\
        \ref{sec:results-exp} & AE (feedforward) & \mat{P} & $64$ \\
        \hline
    \end{tabular}
\end{table}

\subsection{Standard autoencoders}\label{sec:method-network-ae}
\begin{figure}
    \centering
    \begin{subfigure}{\textwidth}
        \includegraphics[width=\textwidth]{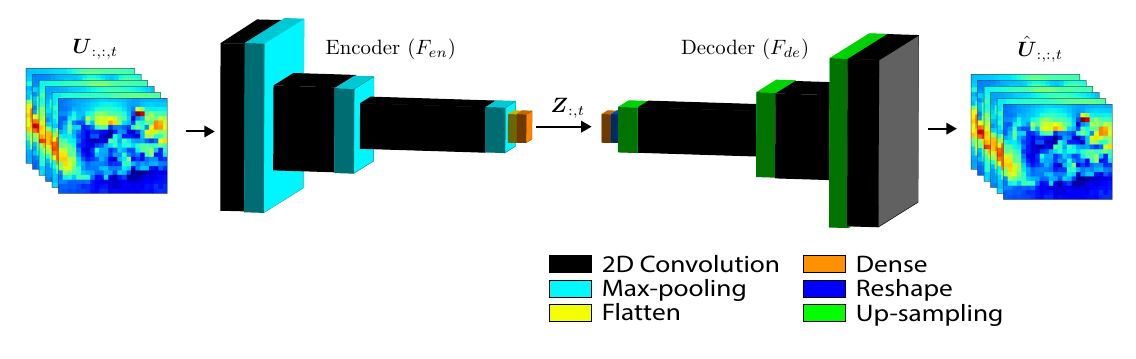}
        \caption{}\label{fig:network-ae-cnn}
    \end{subfigure}
    \begin{subfigure}{\textwidth}
        \includegraphics[width=\textwidth]{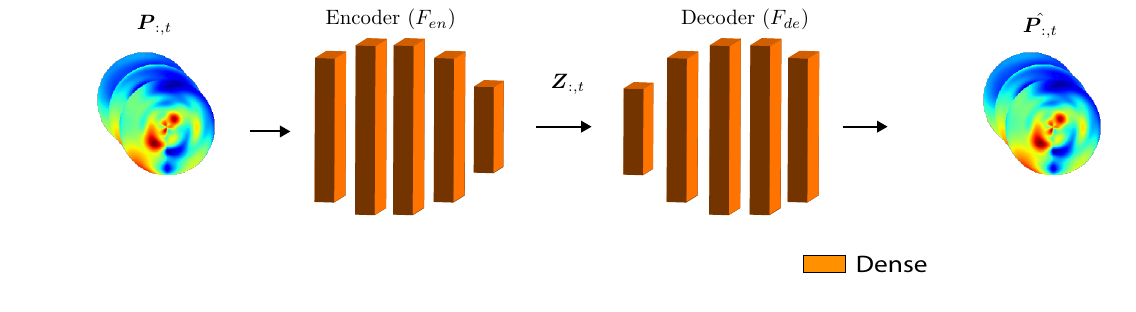}
        \caption{}\label{fig:network-ae-mlp}
    \end{subfigure}
    \caption{
        The schematics of the standard autoencoder (AE). Different AEs are employed for different datasets and tests. The AE architecture and dataset for each test are listed in Table~\ref{tab:test-input-dimension}.         
        a) AE for decomposing the laminar cylinder wake $\mat{U}$ with three convolution layers in both the encoder and the decoder. The hyperparameters are in tables~\ref{tab:appendix-cnn-encoder} and~\ref{tab:appendix-cnn-decoder} in the Appendix.
        b) AE for decomposing the wind-tunnel pressure data $\mat{P}$ with 5 feedforward layers. The input is a flattened vector of measurement taken from all sensors at time $t$. The hyperparameters are given in tables~\ref{tab:appendix-mlp-encoder} and~\ref{tab:appendix-mlp-decoder} in the Appendix.
    }\label{fig:network-ae}
\end{figure}

We refer to a standard autoencoder (AE) as an autoencoder with the structure shown in Figure~\ref{fig:network-ae}, which consists of one encoder $F_{en}$ and one decoder $F_{de}$.
In compact notation, an AE is  
\begin{equation}
    \begin{aligned}
        \vec{Z}_{:,t} &= F_{en}(\vec{Y}_{:,t};\vec{\omega}), \\
        \hat{\vec{Y}}_{:,t} &= F_{de}(\vec{Z}_{:,t};\vec{\omega}). 
        \label{eq:network-ae}
    \end{aligned}    
\end{equation}
Details of the layers are given in the Appendix section~\ref{sec:appendix-layers} (Table~\ref{tab:appendix-cnn-encoder} to~\ref{tab:appendix-mlp-decoder}).
All neural networks that have the convolution encoder and decoder share the same layers and filter size (tables~\ref{tab:appendix-cnn-encoder} and~\ref{tab:appendix-cnn-decoder}).
The feedforward AEs have layers given in tables~\ref{tab:appendix-mlp-encoder}
and~\ref{tab:appendix-mlp-decoder} unless stated otherwise.
For all autoencoders, the activation function `tanh' is used.
All hyperparameters are given in Table~\ref{tab:appendix-hyperparameters}.

\subsection{Mode-decomposing autoencoders (MD-AE)}\label{sec:method-network-mdae}
\begin{figure}
    \centering
    \includegraphics[width=\textwidth]{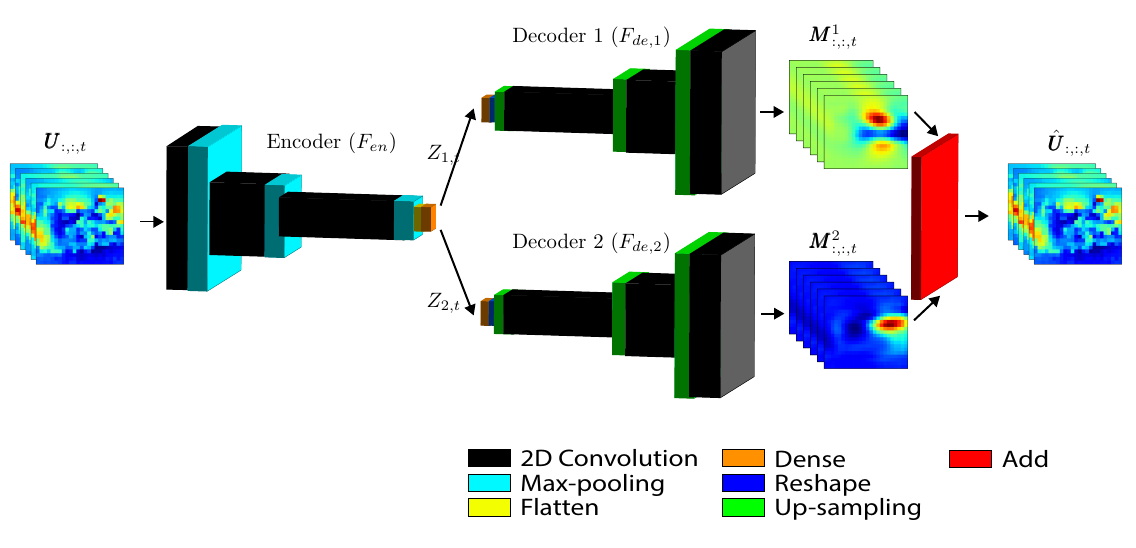}
    \caption{Schematic of the MD-AE \citep{murata2020nonlinear} with two latent variables as an example. Each latent variable is decoded by a decoder to produce a decomposed field. The sum of the decomposed fields $\mat{M}^1$ and $\mat{M}^2$ is the output of the MD-AE.}\label{fig:network-mdae}
\end{figure}

An MD-AE associates one decoder to each latent variable (Figure~\ref{fig:network-mdae}), so we can visualize the effect of each latent variable separately, which makes it easier to interpret \citep{murata2020nonlinear}.
The MD-AE is 
\begin{equation}
    \begin{aligned}
        \vec{Z}_{:,t} &= F_{en}(\vec{Y}_{:,t};\vec{\omega}), \\
        \hat{\vec{Y}}_{:,t} &= \sum_{i=1}^{N_z}
        \underbrace{F^i_{de}(\vec{Z}^i_{:,t};\vec{\omega})}_{\equiv \vec{M}^i_{:,:,t}}
    \end{aligned}
    \label{eq:method-nn-mdae}
\end{equation}
where $\vec{M}^i$ denotes the $i$-th decomposed field of the MD-AE and $\vec{M}^i_{:,:,t}$ is the $i$-th decomposed field at time step $t$, corresponding to the output of the $i$-th decoder.
The MD-AE is only applied to the laminar cylinder wake case (the reasoning behind this decision is explained in section~\ref{sec:results-cylinder}).

\section{The decoder decomposition}\label{sec:method-decoderdecomposition}
The dynamics of POD modes are encapsulated in their time coefficient.
To gain physical insight into the decomposed fields, we propose the decoder decomposition to obtain a relationship between the data time coefficients (Eq. \eqref{eq:method-POD-A}) and the latent variables trained with the same data.
The decoder decomposition is a post-processing method that applies to a trained network.
We define the decoder decomposition for AEs in section~\ref{sec:method-deocoderdecomposition-ae} and for MD-AEs in section~\ref{sec:method-deocoderdecomposition-mdae}.

\subsection{The decoder decomposition for standard autoencoders}\label{sec:method-deocoderdecomposition-ae}

The data modes $\vec{\Phi}^Y$ (the matrix of POD modes of $\mat{Y}$, see Table~\ref{tab:test-input-dimension}) form a basis of the subspace in which the output $\hat{\vec{Y}}_{:,t}$ is approximated as 
\begin{equation}
    \hat{\vec{Y}}_{:,t} = \sum_{k=1}^{N^*} B_{k,t} \vec{\Phi}_{:,k}^Y = \mat{\Phi}^Y \vec{B}_{:,t},
    \label{eq:method-ae-qhat-and-pod}
\end{equation}
with $\mat{B} \in \mathbb{R}^{N^* \times N_t}$ being the temporal coefficients (decoder coefficients).
The decoder coefficient for the data mode $k$ at time step $t$ is $B_{k,t}$.
The matrix form of equation~\eqref{eq:method-ae-qhat-and-pod} is 
\begin{equation}
    \hat{\mat{Y}} = \mat{\Phi}^Y \mat{B}^T.
    \label{eq:method-defineB}
\end{equation}
In a trained network, $\mat{B}$ depends only $\hat{\mat{Y}}$, which depends only on the latent variables $\mat{Z}$. 
Thus, the gradient of the output of the decoder with respect to the latent variables is
\begin{equation}
    \frac{d\hat{\mat{Y}}}{d \vec{Z}} = \mat{\Phi}^Y\frac{d\mat{B}}{d \mat{Z}}.
    \label{eq:decoderdecomposition-take-gradients-matrix}
\end{equation}
The gradient of the decoder coefficients is the sensitivity to changes in the latent variables.
Each element of the gradient matrix $\left(\frac{d\hat{\mat{Y}}}{d \vec{Z}}\right)_{j,i,t}$ is the rate of change of the output at grid point $j$ and time step $t$ ($\hat{Y}_{i,t}$) with respect to the latent variable $Z_{i,t}$.
Because a single decoder coefficient corresponds to a single data mode, which represents a coherent structure, the gradient of the decoder coefficients also reflects the sensitivity of the coherent structure to changes in the latent variables.
Physically, the sensitivity quantifies the relative importance of the different latent variables for decoder coefficients and the data modes they represent.
We quantify the relative importance of the decoder coefficient $\vec{B}_{j,:}$ due to latent variable $\vec{Z}_{i,:}$ by defining the average rate of change 
\begin{equation}
    \epsilon_{i,j} = \frac{
        \int \: \left\lvert \frac{\partial \vec{B}_{j,:}}{\partial \vec{Z}_{i,:}} \right\rvert \: d\vec{Z}_{1,:}\!\dots d\vec{Z}_{N_z}
    }{
        \int \; d\vec{Z}_{1,:}\!\dots d\vec{Z}_{N_z}
    }.
    \label{eq:averaged-rate-of-change}
\end{equation}
The average rate of change of a decoder coefficient due to a latent variable $i$ quantifies its contribution to the data mode $j$ in the output associated with that decoder coefficient (Eq. \ref{eq:method-ae-qhat-and-pod}).
The larger the $\epsilon_{i,j}$, the more important the latent variable $i$ is in representing the data mode $j$ in the output. 
This can be used to rank and select latent variables, as demonstrated in section~\ref{sec:results-exp-select-variables}.
Algorithm~\ref{algo1} shows how the average rate of change is computed.

\RestyleAlgo{ruled}
\SetCommentSty{normalsize}
\begin{algorithm}
\caption{Computing the average rate of change}\label{algo1}
\SetKwFunction{std}{$\sigma$} 
\SetKwInOut{Input}{Input}
\SetKwInOut{Output}{Output}
\SetKwInOut{Require}{Require}

\Input{
    $F_{de}$ - trained decoder\\
    $\Delta z$ - constant step size
}
\Output{$\mat{\epsilon}$ - average rate of change}

\For{$i \gets 1,\dots, N_z$}{
    $\frac{d \hat{\vec{Y}}}{d \vec{Z}}_{i,:} \gets  \frac{1}{\std(\vec{Z_{i,:}})} \frac{F_{de}(\vec{Z}_{i,:} + \Delta z) - F_{de}(\vec{Z}_{i,:} - \Delta z)}{2 \Delta z}$ 
    \tcp*{approximate derivatives \& normalize}
}

\text{Collect }$\frac{d \hat{\mat{Y}}}{d \mat{Z}}$;

$\frac{d \mat{B}}{d \mat{Z}} \gets {\mat{\Phi}^Y}^T \frac{d \hat{\mat{Y}}}{d \mat{Z}}$;

$\mat{\epsilon}^T \gets \frac{1}{N_t}\sum^{N_t}_{t=1}\left|\frac{d \mat{B}}{d \mat{Z}}\right|_{:,:,t}$ \tcp*{average rate of change}
\KwRet $\mat\epsilon$

\end{algorithm}

\subsection{Decoder decomposition for MD-AEs}\label{sec:method-deocoderdecomposition-mdae}

In an MD-AE, the decomposed fields $\mat{M}^i$ is decomposed into 
\begin{equation}
     \mat{M}^i = \mat{\Phi}^Y {\mat{B}^i}^T,  \label{eq:def_Bi}
\end{equation}
where $\mat{B}^i$ is the matrix of the $i$-th decoder coefficients of the $i$-th decomposed field.
Combined with equation~\eqref{eq:method-nn-mdae}, the output of the MD-AE becomes
\begin{equation}
    \hat{\mat{Y}} = \mat{\Phi}^Y \sum_{i=1}^{N_z}{\mat{B}^i}^T.
\end{equation}
Equations~\eqref{eq:decoderdecomposition-take-gradients-matrix} to~\eqref{eq:averaged-rate-of-change} still apply as they are.

\section{Decomposition of the unsteady laminar wake}\label{sec:results-cylinder}

In this section, we analyse the latent space of autoencoders trained with the laminar cylinder wake dataset (section~\ref{sec:data-cylinder}).
We first decompose the dataset with an MD-AE and apply the decoder decomposition in section~\ref{sec:result-cylinder-mdae}, which serves as a benchmark to compare against the literature.
We then apply the decoder decomposition to interpret the latent variables of AEs in section~\ref{sec:result-cylinder-ae}.

\subsection{The latent space of a mode-decomposing autoencoder}\label{sec:result-cylinder-mdae}
\begin{figure}
    \centering
    \includegraphics[width=0.75\textwidth]{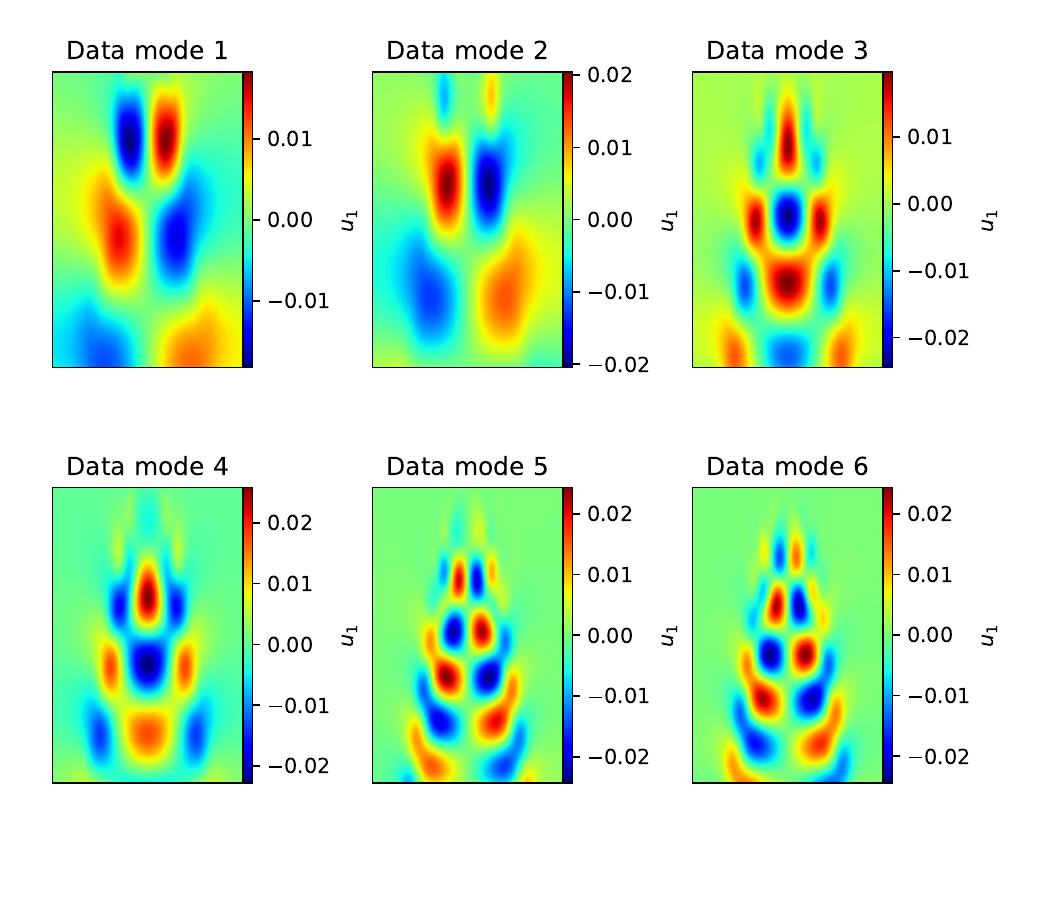}
    \caption{The first six data modes of the unsteady wake behind a cylinder dataset $\mat{\Phi}^{lam}_{:,1},\dots,\mat{\Phi}^{lam}_{:,6}$.}\label{fig:cylinder-data-pod}
\end{figure}

\begin{figure}
    \centering
    \begin{subfigure}[b]{0.37\textwidth}
        \includegraphics[width=\textwidth]{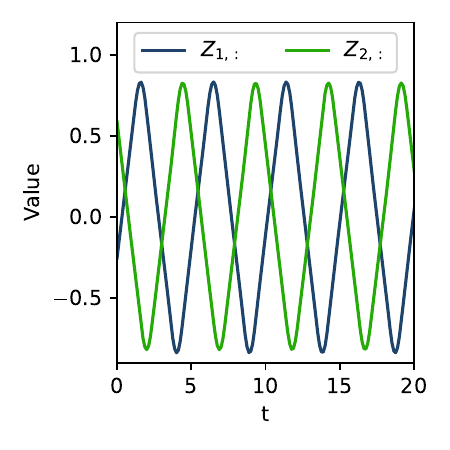}
        \caption{}\label{fig:cylinder-mdae-summarya}
    \end{subfigure}
    \begin{subfigure}[b]{0.50\textwidth}
        \includegraphics[width=\textwidth]{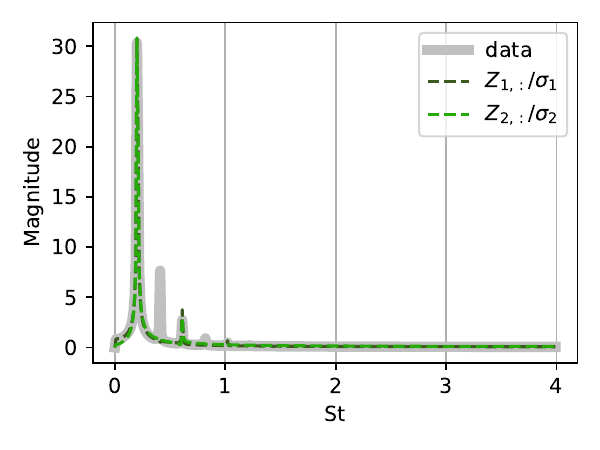}
        \caption{}\label{fig:cylinder-mdae-summaryb}
    \end{subfigure}
    \begin{subfigure}{0.85\textwidth}
        \includegraphics[width=\textwidth]{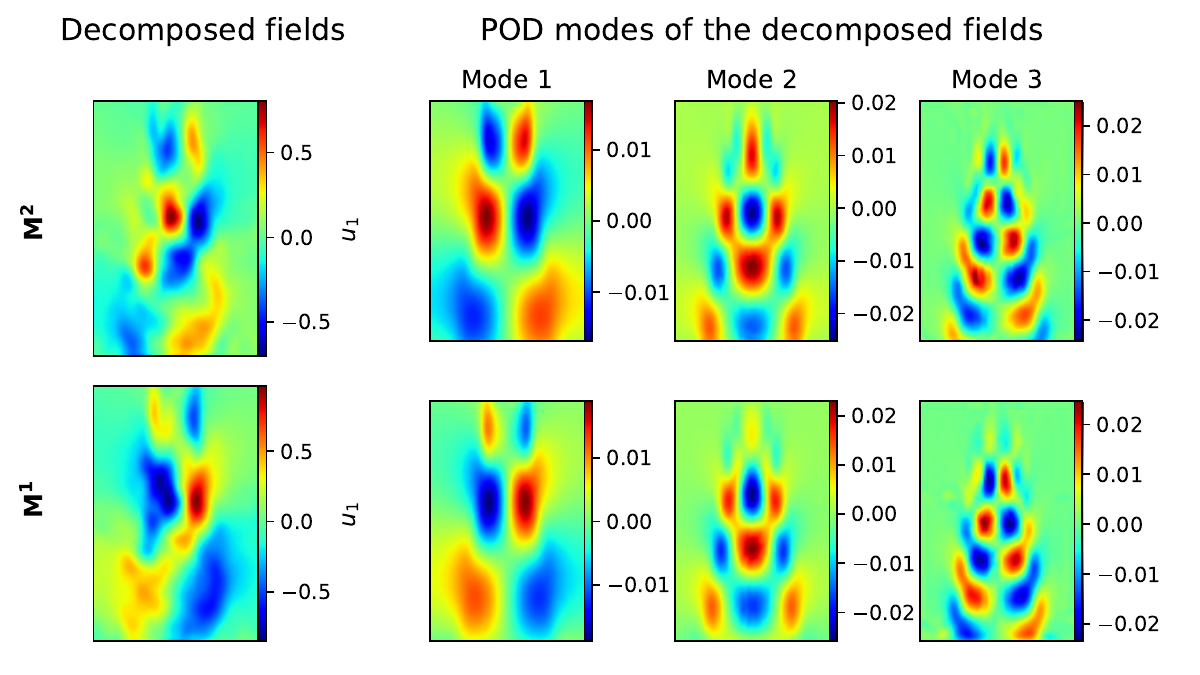}
        \caption{}\label{fig:cylinder-mdae-summaryc}
    \end{subfigure}
    \caption{
        Results of training the MD-AE on the laminar wake dataset with a latent dimension of 2.
        a) Both latent variables are periodic in time.
        b) Frequency spectrum of the latent variables, normalised by their standard deviation, compared with the frequency of the data. The latent variables both contain the vortex shedding frequency and the second harmonic of the vortex shedding frequencies.
        c) POD modes of the decomposed fields. The POD modes of decomposed field 1 are similar to data mode 1,3 and 6 and the POD modes of decomposed field 2 are similar to data mode 2,4 and 5.}\label{fig:cylinder-mdae-summary}
\end{figure}

We decompose the unsteady wake dataset, described in section~\ref{sec:data-cylinder}, with an MD-AE with $N_z=2$ (Figure~\ref{fig:cylinder-mdae-summary}) to verify the decoder decomposition.
The reconstruction loss, measured with MSE (Eq.~\eqref{eq:mse}), is $2.6 \times 10^{-5}$, showing that the output $\hat{\mat{U}}$ is an accurate approximation of $\mat{U}$. 
Figure~\ref{fig:cylinder-mdae-summarya} shows that the latent variables have the same periodic behaviour as the data time coefficients, matching the results of \citep{murata2020nonlinear}.
Figure~\ref{fig:cylinder-data-pod} shows the data modes of the laminar dataset.
By applying POD to the decomposed fields and by inspection, the decomposed field 1 contains the data modes 1, 3 and 6, and the decomposed field 2 contains the data modes 2,4 and 5 (Figure~\ref{fig:cylinder-mdae-summaryc}).
Figure~\ref{fig:cylinder-mdae-summaryb} shows that the two latent variables have the same frequency spectrum, with peaks at $St= 0.23$ and $0.69$, corresponding to the vortex shedding frequency and its second harmonic.
Because the decomposed field 1 contains a POD mode similar to data mode 1 and the decomposed field 2 contains a similar mode to data mode 2, and the two latent variables have an identical frequency spectrum, are periodic and out of phase, we conclude that the two latent variables form a pair similar to the data time coefficients. 
However, the latent variables have two frequency peaks instead of a singular peak of the first two data time coefficients, showing that the latent variables contain nonlinear temporal information.\\ 

\begin{figure}
    \centering
    \FIG{
        \includegraphics[width=0.9\textwidth]{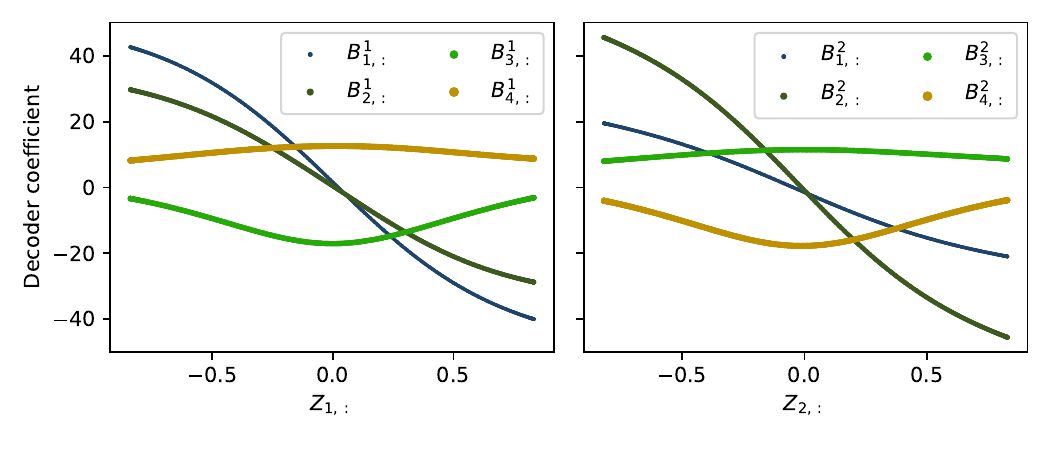}
    }{
        \caption{
            Results from the MD-AE trained with the laminar wake dataset with 2 latent variables.
            The first four decoder coefficients plotted against the latent variables. 
            Both latent variables affect the magnitude of the first four POD modes of the data in the output of the MD-AE.} 
    }\label{fig:cylinder-mdae-decodercoeff}
\end{figure}

We apply the decoder decomposition to the MD-AE and plot the first four decoder coefficients for the two decomposed fields in Figure~\ref{fig:cylinder-mdae-decodercoeff}.
None of the first four decoder coefficients are constant in Figure~\ref{fig:cylinder-mdae-decodercoeff}, meaning that the first four data modes all contribute toward both decomposed fields. 
The relative frequencies of the latent variables and the decoder coefficients can also be deduced from Figure~\ref{fig:cylinder-mdae-decodercoeff}. 
As the latent variables change from $-1$ to $1$, decoder coefficients 1 and 2 in both decomposed fields complete a quarter of a sine wave, but the decoder coefficients 3 and 4 complete half a sine wave, meaning that decoder coefficients 3 and 4 oscillate at double the frequency of decoder coefficients 1 and 2.

For a more quantitative measure of how the data modes are split into the decomposed fields of the MD-AE, we calculate the equivalent energy \citep{kneer2021symmetry-aware} for each decomposed field separately.
The matrix of equivalent energy for the $i$-th decomposed field and the data modes of the laminar flow, $\hat{\mat{\Lambda}}^i$, is given by
\begin{equation}
    \hat{\mat{\Lambda}}^i = \frac{1}{N_t-1}\mat{B}^i {\mat{B}^i}^T,
    \label{eq:equivalent_pca}
\end{equation}
which is interpreted as the variance of the $i$-th decomposed field, $\mat{M}^i$, projected onto the data modes.
The equivalent energy of the $i$-th decomposed field and the $k$-th data mode is $\hat{\Lambda}^i_{kk}$.
As $\mat{M}^i$ approaches the first data POD decomposed field ($\vec{\Phi}^{lam}_{:,1} \boldsymbol{A}_{i,:}^T$), $\hat{\Lambda}^i_{11}$ approaches $\Lambda_{11}$ while $\hat{\Lambda}^i_{kk}$, where $k\neq1$, approaches $0$.

\begin{figure}
    \centering
    \FIG{
        \includegraphics[width=0.68\textwidth]{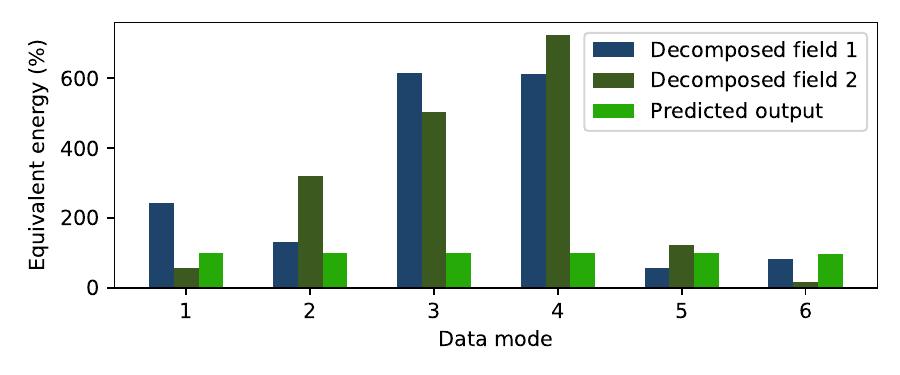}
    }{
        \caption{
            Results from the MD-AE trained with the laminar wake dataset with 2 latent variables.
            The equivalent energy of the two decomposed fields of the MD-AE, showing the first four POD modes of the data. All POD modes of data are present in both decomposed fields.}
    }\label{fig:cylinder-mdae-equivalentenergy}
\end{figure}

Figure~\ref{fig:cylinder-mdae-equivalentenergy} shows the equivalent energy of the first six data modes in the two decomposed fields and in the predicted output $\hat{\mat{U}}$.
The equivalent energies of the predicted output for all data modes are approximately $100\%$, meaning that the output $\hat{\mat{U}}$ is an accurate approximation of $\mat{U}$.
which also shows that the MD-AE does not fully separate data modes into different decomposed fields.

In summary, the decoder decomposition applied to an MD-AE~\citep{murata2020nonlinear} leads to a detailed interpretation of the decomposed fields. 
By computing the equivalent energy, we also show that, although the MD-AE works well for visualisation, the decomposed fields are entangled.

\subsection{The latent space of standard autoencoders}\label{sec:result-cylinder-ae}

Unlike the MD-AE, an AE has no decomposed fields to visualize.
We apply the decoder decomposition to AEs trained with the unsteady wake dataset to connect the data time coefficients to the latent variables.
Since the frequency content of the laminar, periodic wake of a cylinder and the data modes are well known \citep{m.m.zdravkovich1997FlowCircularCylinders}, we use the interpretation of the data modes and time coefficients to provide an interpretation of the latent variables.

\begin{figure}
    \centering
\includegraphics[width=0.6\textwidth]{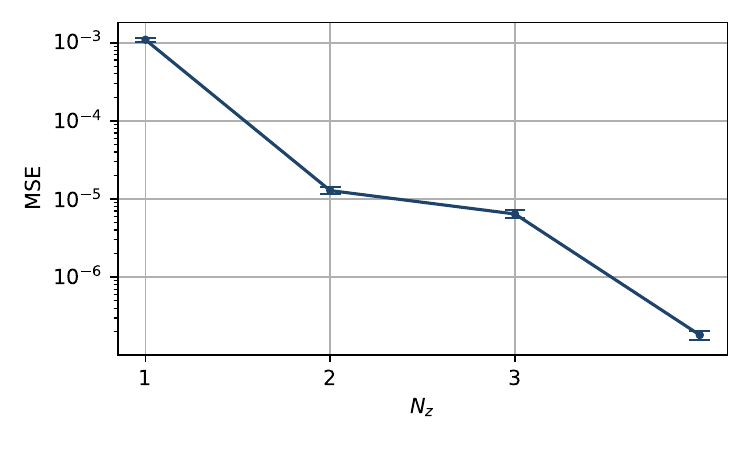}
    \caption{The MSE of AEs trained with the unsteady wake dataset with different numbers of latent variables averaged over five repeats each. The error bars represent the standard deviation of the repeats. At $N_z=2$ the MSE is approximately $1.3 \times 10^{5}$.
    }\label{fig:cylinder-ae-convergence}
\end{figure}

Figure~\ref{fig:cylinder-ae-convergence} shows the loss of AEs trained with the unsteady wake dataset using 1, 2, 3 and 4 latent variables. 
The first plateau of loss occurs at $N_z=2$ before the loss decreases again at $N_z=4$.
Similar behaviour is also observed by \citep{csala2022ComparingDifferentNonlinear}.
The AEs with two latent variables achieve a significantly smaller loss than AEs with a single latent variable but the difference in loss between using two and three latent variables is small.
The laminar wake $\mat{U}$ has periodic underlying dynamics and thus can be described by a single variable in polar coordinates (the angle) or two variables in Cartesian coordinates. 
However, the AEs do not find the best representation when forced to use only one latent variable due to their inability to perform Cartesian to polar transformation.
Polar to Cartesian relies on the $\sin$ function, which is a many-to-one function and can be learned by neural networks either through approximation with any activation function or by using $sin$ as the activation function \citep{wong2022LearningSinusoidalSpaces}.
However, the Cartesian to polar transformation relies on $\arcsin$, which is not defined for all inputs.
The Cartesian to polar transformation is a known difficulty in machine learning, with benchmark problems such as the two-spirals problem built around it \citep{fahlman1989CascadeCorrelationLearningArchitecture,liu2018IntriguingFailingConvolutional}, this is consistent with our observations (Figure~\ref{fig:cylinder-ae-convergence}).
Therefore, we analyse an AE with $N_z=2$ with decoder decomposition to obtain physical interpretation in the current section, even though $N_z=1$ should be the theoretical minimum representation of the laminar wake.

\begin{figure}
    \centering
    \begin{subfigure}{0.28\textwidth}
        \includegraphics[width=\textwidth]{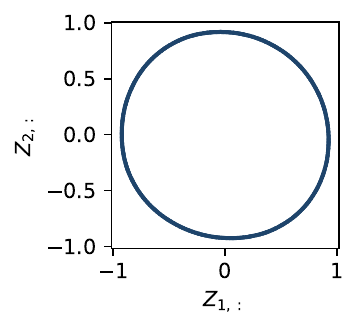}
        \caption{}\label{fig:cylinder-ae2-latentbasica}
    \end{subfigure}
    \begin{subfigure}{0.28\textwidth}
        \includegraphics[width=\textwidth]{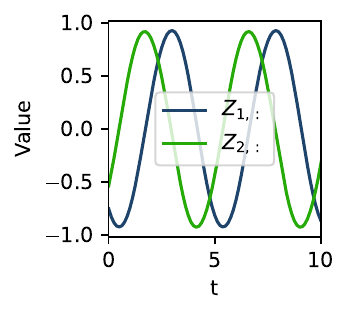}
        \caption{}\label{fig:cylinder-ae2-latentbasicb}
    \end{subfigure}
    \begin{subfigure}{0.3\textwidth}
        \includegraphics[width=\textwidth]{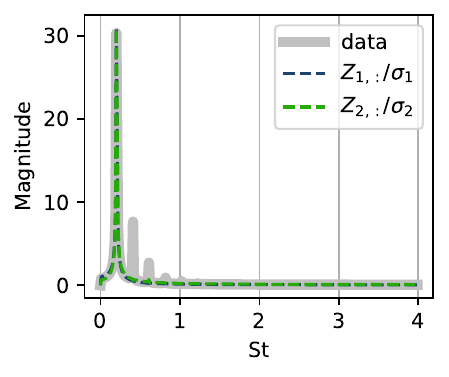}
        \caption{}\label{fig:cylinder-ae2-latentbasicc}
    \end{subfigure}
    \caption{
        Results from the AE trained with the laminar wake dataset with two latent variables.
        a) Phase plot of two latent variables. The unit circle indicates harmonic oscillations.
        b) Time trace of the two latent variables. The latent variables are periodic and $90^\circ$ out of phase, behaving the same as the first two data time coefficients.
        c) The frequency spectrum of the latent variables, compared to the frequency in the dataset. The variables are normalized by their standard deviation, $\sigma$, before applying the Fourier Transform for visualization. 
    }\label{fig:cylinder-ae2-latentbasic}
\end{figure}

The phase plot of the two latent variables is a unit circle (Figure~\ref{fig:cylinder-ae2-latentbasica}), which indicates that the latent variables of an AE are periodic; the same phase plot also describes the first two data time coefficients.
The same periodic behaviour also shows in the time trace of the latent variables (Figure~\ref{fig:cylinder-ae2-latentbasicb}).
Figure~\ref{fig:cylinder-ae2-latentbasicc} compares the frequency spectrum of the latent variables and the data. 
We find that both latent variables oscillate at the vortex shedding frequency, which is the lowest frequency in the dataset (Figure~\ref{fig:cylinder-data-pod-results}, right panel).
The higher harmonic frequencies are not included in the latent variables because they are functions of the vortex-shedding frequencies, and the decoders can represent the spatial patterns of multiple data modes.
The AE has produced latent variables which have a singular peak in the frequency spectrum, less than the frequency spectrum of the latent variables of the MD-AE, without sacrificing accuracy, 
showing that AEs are more expressive given the same number of latent variables than MD-AEs.

\begin{figure}
    \centering
    \includegraphics[width=0.7\textwidth]{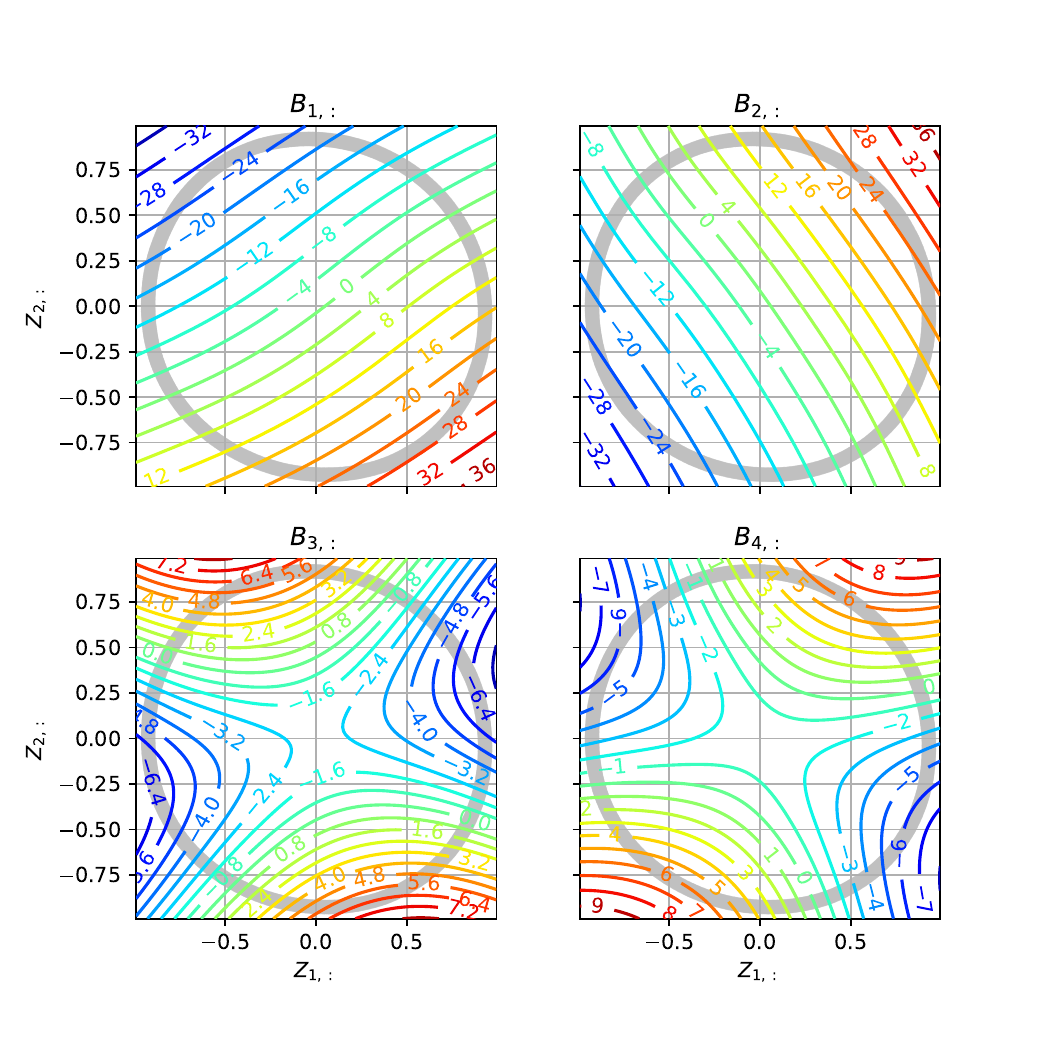}
    \caption{
        Results from the AE trained with the laminar wake dataset with two latent variables.
        Contour plot of decoder coefficients of an AE with two latent variables, trained with the unsteady wake dataset.
        The contours show the values of the decoder coefficients.
        The grey circle labels the values of the latent variables observed during training, which shows the dynamics of the dataset.
    }\label{fig:cylinder-ae2-contour}
\end{figure}

Figure~\ref{fig:cylinder-ae2-contour} shows the sensitivity of the decoder coefficients, $\vec{B}_{1,:} , \dots, \vec{B}_{4,:}$ to the two latent variables.
The contours are the values of the decoder coefficients, shown for all values of the latent variables allowed by the nonlinear activation function.
The dynamics of the dataset are shown as the grey circle, which shows the values of the latent variables that are observed during training.

The decoder coefficients $\vec{B}_{3,:}$ and $\vec{B}_{4,:}$ have double the frequency of $\vec{B}_{1,:}$ and $\vec{B}_{2,:}$, consistent with the characteristics of the first four data modes.
At $\mat{Z} = \mat{0}$, all decoder coefficients have the value $0$ because the network has no bias.
Figure~\ref{fig:cylinder-ae2-contour} provides a qualitative overview of the sensitivity of the data modes to the latent variables.

\begin{figure}
    \centering
    \includegraphics[width=0.5\textwidth]{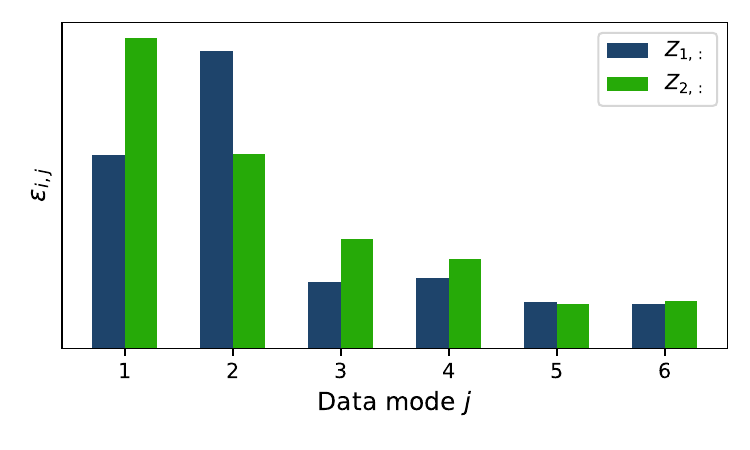}
    \caption{
        Results from the AE trained with the laminar wake dataset with 2 latent variables.
        The average rate of change of the decoder coefficients.
        }\label{fig:cylinder-ae2-rateofchange}
\end{figure}

Because the decoder output $\hat{\mat{U}}$ depends on multiple latent variables, the equivalent energy (Eq.~\eqref{eq:equivalent_pca}) cannot be used to identify the contribution of each latent variable.
We use the average rate of change (Eq.~\eqref{eq:averaged-rate-of-change}) to quantify the contribution of individual latent variables of the AE (Figure~\ref{fig:cylinder-ae2-rateofchange}).
Amongst the first four data modes, latent variable $\vec{Z}_{1,:}$ contributes most to the presence of the second data mode $\vec{\Phi}_{:,2}^{lam}$ in the output $\hat{\mat{U}}$, while $\vec{Z}_{2,:}$ mainly contributes towards the presence of the other three data modes.
The first 6 data modes depend on both latent variables, so no latent variable can be removed without affecting the dynamics of the output.
However, the average rate of change suggests that the smaller-scale energy-containing structures between the vortex streets \citep{m.m.zdravkovich1997FlowCircularCylinders}, namely the pair comprised of data mode 3 and 4, are mainly represented by $\vec{Z}_{2,:}$. 
Therefore, $\vec{Z}_{2,:}$ should be used if the representation of data modes 3 and 4 in the output is required.

By comparing the frequency spectrum of the latent variables of the AE and the MD-AE (section~\ref{sec:result-cylinder-mdae}), both with $N_z=2$, we find that the AE has learned a more accurate representation of the data than the MD-AE because the latent variables have the minimum number of frequency peaks needed to represent the data.
The frequency of the two latent variables coincides with the lowest frequency in the frequency spectrum of the data, meaning that the data modes with higher frequencies are added by the decoder.
By introducing the average rate of change,  
we show that both latent variables are equally important in representing the vortex shedding, but the first harmonic is contained mainly in one of the latent variables in this case.
Because the training of an AE depends on random seeds, the physical meaning of the latent variables may change for different cases. 
During training, we noticed that the dimension of latent space has a significant impact on the physical interpretation of the latent variables. 
In-depth comments with example AEs trained with $N_z=1$ and $N_z=3$ are found in Appendix~\ref{sec:result-cylinder-ae-wrongnz}.

\section{Autoencoder decomposition of the turbulent wake from wind-tunnel experiments}\label{sec:results-exp}
We analyse the AEs to decompose the wind-tunnel wake dataset $\mat{P}$ (section~\ref{sec:data-exp}).
We show that, for a turbulent dataset, the size of the network becomes an additional limiting factor for the interpretability of AEs in section~\ref{sec:results-exp-small-nz}.
We demonstrate how to select latent variables to single out a particular flow structure in section~\ref{sec:results-exp-select-variables}.

\subsection{The limiting factor is network size for turbulence}\label{sec:results-exp-small-nz}

\begin{figure}
    \centering
    \includegraphics[width=\textwidth]{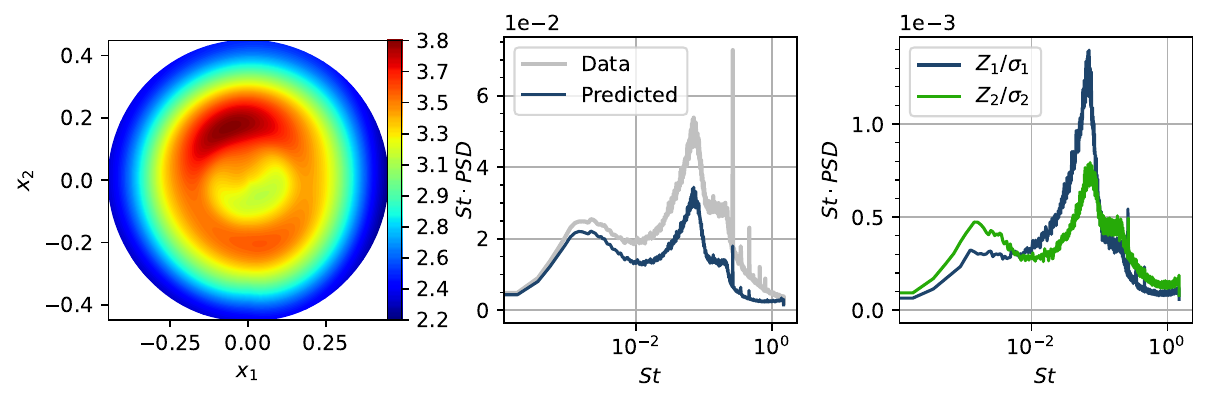}
    \caption{
        The AE trained with the wind-tunnel dataset with two latent variables.
        Left: predicted RMS pressure.
        Centre: the premultiplied overall PSD of the prediction. 
        Right: the premultiplied PSD of the latent variables of the AE trained with two latent variables, normalized by their standard deviation.
        }\label{fig:exp2-pred}
\end{figure}

\begin{figure}
    \centering
    \includegraphics[width=\textwidth]{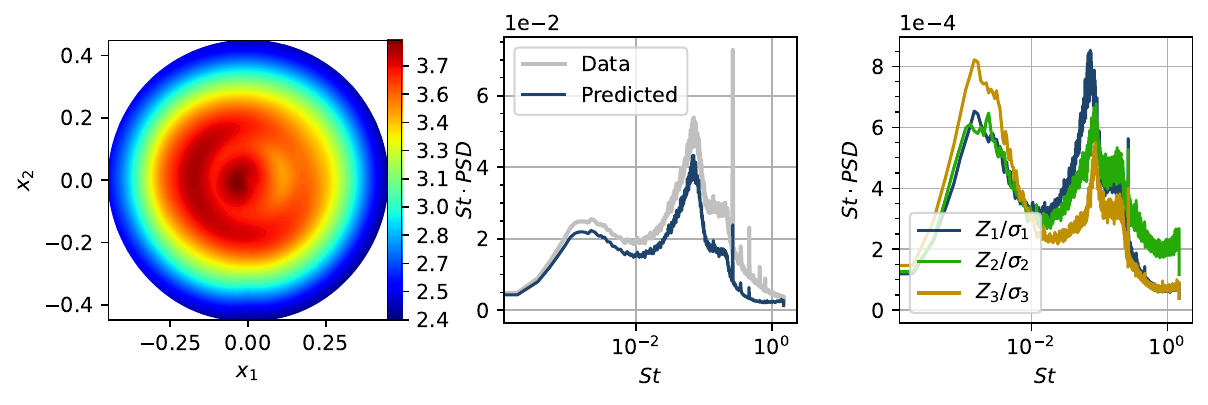}
    \caption{
        Same as Figure~\ref{fig:exp2-pred} with three latent variables.
    }\label{fig:exp3-pred}
\end{figure}

At least two latent variables are needed to represent flows with periodicity (section~\ref{sec:result-cylinder-ae}). 
Because the turbulent wake of the axisymmetric body contains periodic behaviours such as vortex shedding \citep{rowand.brackston2017FeedbackControlThreeDimensional}, we first train an AE with two latent variables on the wind-tunnel dataset (feedforward, see Table~\ref{tab:test-input-dimension}).
Figure~\ref{fig:exp2-pred} (left, centre panels) shows the RMS pressure and the frequency content of the predicted pressure. 
The frequency peaks of the reconstructed pressure field match the data, but the RMS pressure and the magnitude of the PSD differ from the reference pressure field.
We investigate the effect of increasing the latent variables from two to three.
The prediction made by the AE with $N_z=3$ is more accurate in both the RMS pressure and frequency content (Figure~\ref{fig:exp3-pred}, left panel, centre panels) compared to the prediction by the AE with $N_z=2$.
By increasing the number of the latent variables from two to three, the prominent frequencies of the prediction remain unchanged, but the magnitude is closer to the reference data.
By increasing the dimension of the latent space from $N_z=2$ to $N_z=3$, the prediction has improved without changing any frequency-related behaviour. 
This suggests that the benefit of increasing the latent dimension comes from increasing the size of the decoder, thus improving the decoder's ability to express spatial patterns.

\begin{figure}
    \centering
    \includegraphics[width=0.9\textwidth]{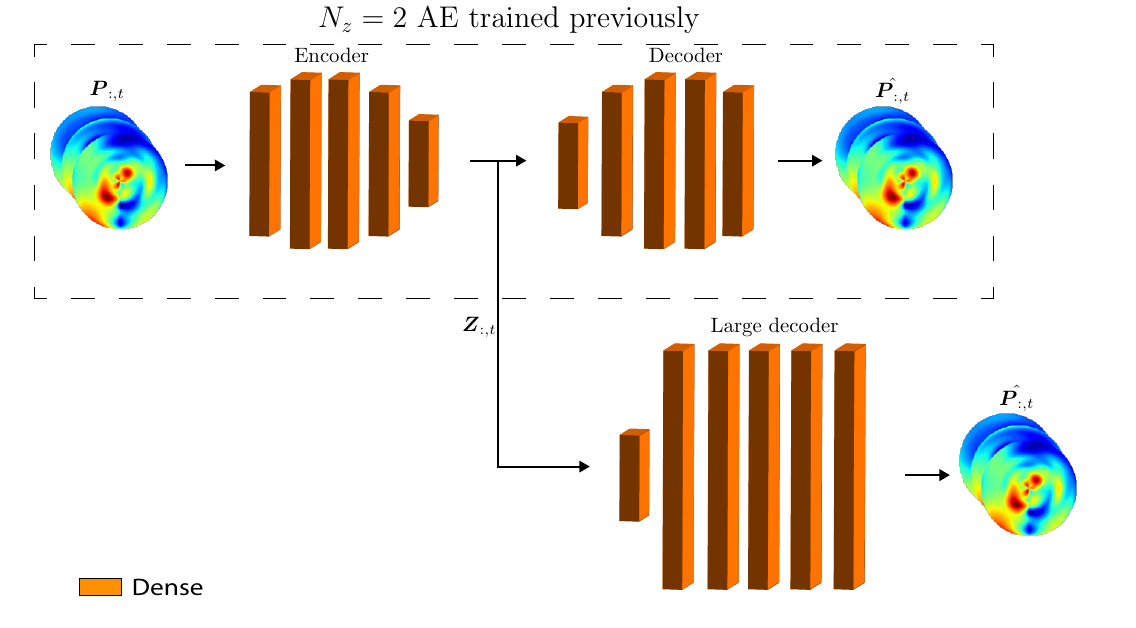}
    \caption{
        The schematics and training process of the decoder-only network. 
        The decoder has 99\% more trainable parameters than the decoder in the $N_z=2$ AE discussed in section~\ref{sec:results-exp-small-nz}.
    }\label{fig:decoder-only-schematics}
\end{figure}

To further understand the cause of the improved prediction, we train a large decoder-only network (Figure~\ref{fig:decoder-only-schematics}) with the latent variables obtained with the previous $N_z=2$ AE. (The details of the large decoder are given in Table~\ref{tab:appendix-large-decoder}.) 
The large decoder has approximately twice the number of trainable parameters than the decoder in the previous $N_z=2$ AE.\@
The input to the large decoder is the latent variables obtained from the trained $N_z=2$ AE, which have already been analysed in Figure~\ref{fig:exp2-pred} (right panel); the output of the large decoder approximates the reference fluctuating pressure $\mat{P}$.

\begin{figure}
    \centering
    \includegraphics[width=0.55\textwidth]{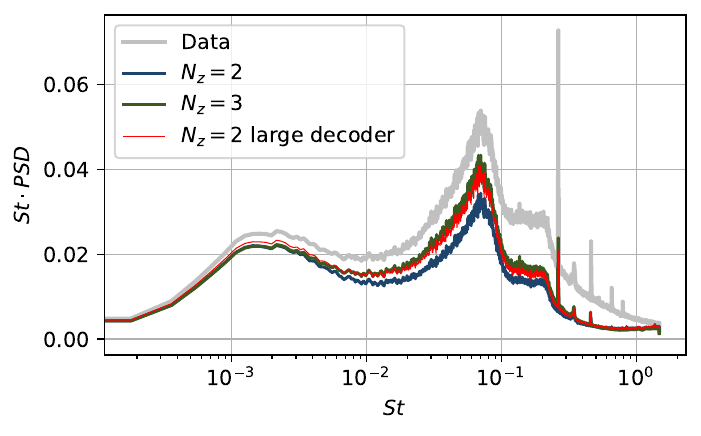}
    \caption{
        The premultiplied PSD of the predicted pressure from the AE with two and three latent variables, and from the large decoder-only network, trained with the wind-tunnel dataset.}\label{fig:decoder-only-pred}
\end{figure}

The PSD of the large decoder-only network overlaps with the PSD of the $N_z=3$ smaller-sized AE (Figure~\ref{fig:decoder-only-pred}), both are closer to the PSD of the reference data than the smaller $N_z=2$ AE.\@
When the latent variables remain unchanged, the large decoder-only $N_z=2$ network can make predictions that match the predictions made by a smaller dimension $N_z=3$ AE.\@
By comparing the results of the smaller size $N_z=2$ and $N_z=3$ AEs to the large decoder-only network with identical latent variable to the $N_z=2$ AE, we find that using a larger decoder has the same effect as increasing the number of latent variables.
More importantly, the frequency peaks in the latent variables and predictions remain the same for all three networks.
Therefore, we conclude that the critical factor for an accurate prediction using the wind-tunnel dataset, which contains multiple spatiotemporal scales, is the size of the decoder.
In an AE, the weights are responsible for representing the spatial patterns in the flow, and the latent variables represent the time-dependent behaviours.
We found that the optimal dimension of the latent space is a function of both of the underlying manifold's dimension~\citep{magril.2022InterpretabilityProperLatent,doan2023ConvolutionalAutoencoderSpatiotemporal,racca2023predicting} and the size of the decoder.

\subsection{Filtering unwanted latent variables}\label{sec:results-exp-select-variables}

\begin{figure}
    \centering
    \includegraphics[width=0.8\textwidth]{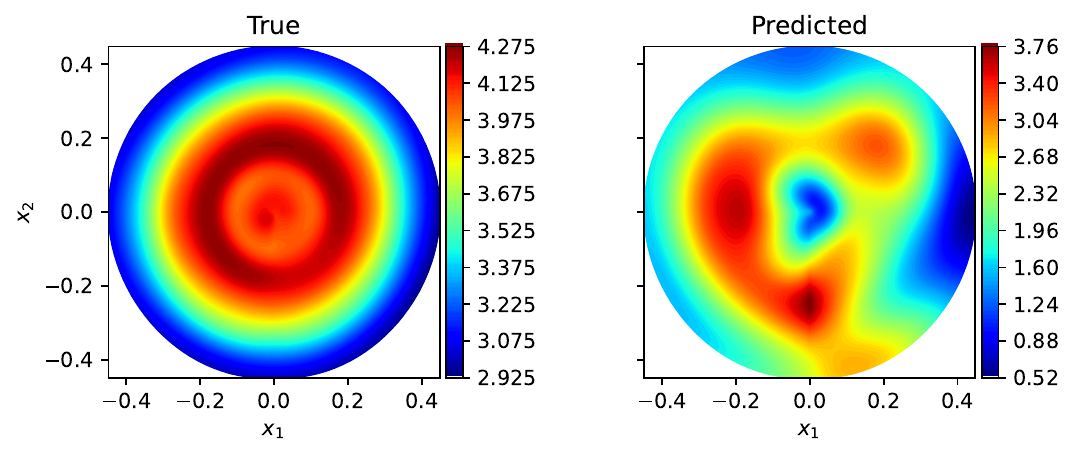}
    \caption{
        Prediction obtained with the AE with 28 latent variables from the filtered latent variables $\mat{Z}_f$ and the reference data.
        }\label{fig:exp28-filtered-rms}
\end{figure}

\begin{figure}
    \centering
    \includegraphics[width=0.6\textwidth]{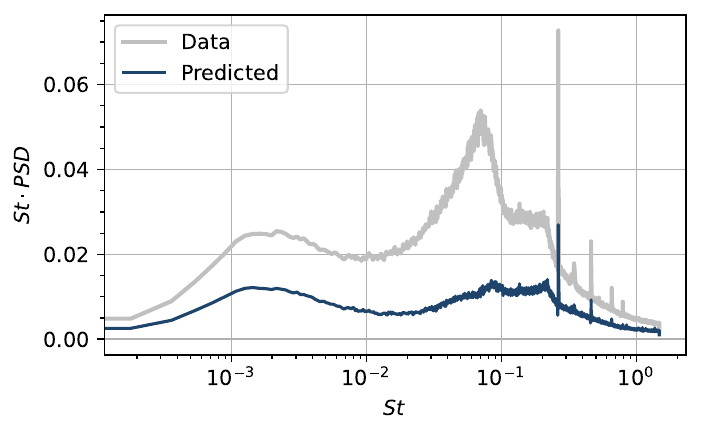}
    \caption{
        Premultiplied PSD of the prediction obtained with the AE with 28 latent variables from the filtered latent variables. All peaks (except for $St \approx 0.2$) are filtered. }
    \label{fig:exp28-filtered-psd}
\end{figure}

Section~\ref{sec:results-exp-small-nz} shows that changing both the dimension of the latent space, $N_z$, and the size of the AEs have a similar effect for the turbulent datasets.
In this section, we apply the proposed decoder decomposition to filter out unwanted latent variables.
\begin{figure}
    \centering
    \includegraphics[width=0.55\textwidth]{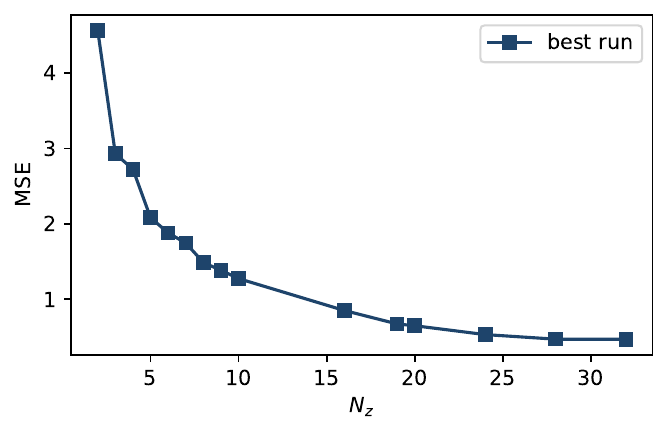}
    \caption{AE with different numbers of latent variables and the same hyperparameters, trained with the wind-tunnel dataset. 
    The loss stops decreasing at $N_z\approx28$.
    }\label{fig:exp-convergence}
\end{figure}
When training on the turbulent pressure dataset, $\mat{P}$, we find that the MSE is sufficiently small at $N_z\approx28$ (Figure~\ref{fig:exp-convergence}).
Therefore, we employ the AE with 28 latent variables. 
We filter the latent variables to enhance or reduce the importance of certain data modes among other data modes in the output of the AEs, similar to selecting data modes for reconstructing only a particular coherent structure.
\begin{figure}
    \centering
    \includegraphics[width=0.8\textwidth]{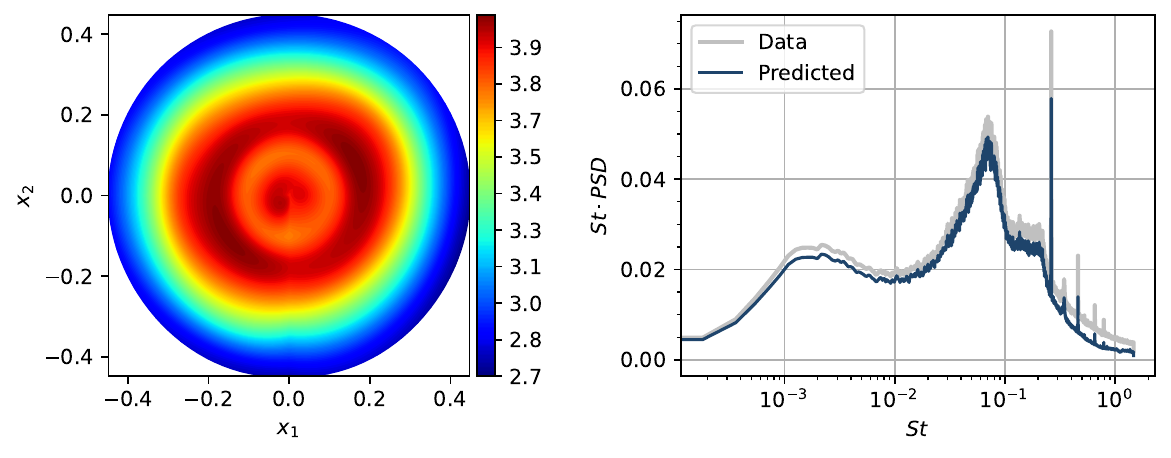}
    \caption{RMS predicted pressure and premultiplied PSD of the prediction of the AE trained with the wind-tunnel dataset with 28 latent variables.
    The network attains an MSE equivalent to the reconstruction with 30 POD modes.
    }\label{fig:exp28-pred}
\end{figure}
\begin{figure}
    \centering
    \begin{subfigure}{0.47\textwidth}
        \includegraphics[width=\textwidth]{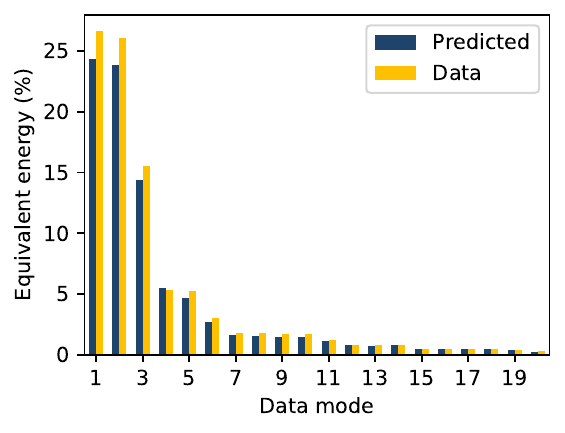}
        \caption{}\label{fig:exp28-control-equivalent-energy}
    \end{subfigure}
    \begin{subfigure}{0.49\textwidth}
        \includegraphics[width=\textwidth]{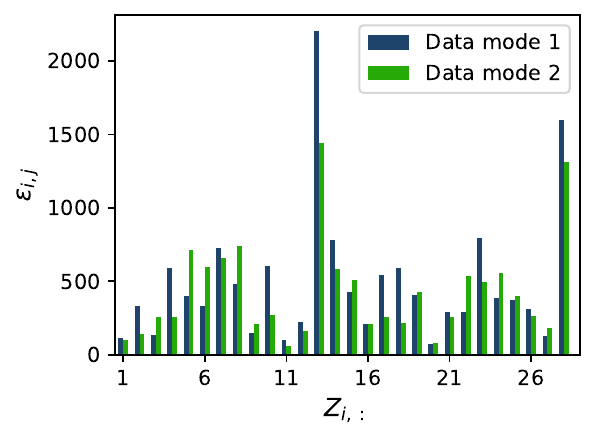}
        \caption{}\label{fig:exp28-rate-of-change}   
    \end{subfigure}
    \caption{
        The AE trained with the wind-tunnel dataset with 28 latent variables.
        a) Equivalent energy of the predicted pressure compared to the POD eigenvalues of the reference data.
        b) The average rate of change of decoder coefficients 1 and 2 due to latent variables, normalized by the standard deviation of the latent variables. The decoder coefficients 1 and 2 are direct analogies of the data POD time coefficients 1 and 2, which are associated with vortex shedding.
    }
\end{figure}
Figure~\ref{fig:exp28-pred} shows the results from the AE with 28 latent variables. 
The RMS predicted pressure is in agreement with the reference RMS pressure, and the prediction has PSD that approximates the data's PSD in terms of both magnitude and frequency content.
To understand which data modes are present in the prediction, we define the equivalent energy for an AE with a POD weight matrix, $\mat{W}_p$ (Eq.~\ref{eq:method-POD-C}), as
\begin{equation}
    \hat{\mat{\Lambda}} = \frac{1}{N_t-1} {\mat{\Phi}^Y}^T \left( ( \mat{W}_p {\mat{\Phi}}^Y \mat{B}^T )   ( \mat{B} {\mat{\Phi}^Y}^T \mat{W}_p^T )  \right)\mat{\Phi}^Y
\end{equation}
from equation~\ref{eq:equivalent_pca}~\citep{kneer2021symmetry-aware}. 
The matrix
$\mat{\Phi}^Y=\mat{\Phi}^{\sym{exp}}$ contains the data modes. 
Figure~\ref{fig:exp28-control-equivalent-energy} compares the equivalent energy of the predicted pressure from the AE with $N_z=28$ with the POD eigenvalues of the reference data.
The prediction contains a similar amount of energy to the reference data for the more energetic coherent structures identified by POD.
We focus our analysis on data modes 1 and 2, which represent vortex shedding.
Figure~\ref{fig:exp28-rate-of-change} shows how each latent variable affects the decoder coefficients 1 and 2 in the prediction, which are associated with POD modes 1 and 2 in the prediction.
We compute the average rate of change $\epsilon_{i,j}$ (equation~\eqref{eq:averaged-rate-of-change}) and focus the analysis on data modes 1 and 2, which represent vortex shedding.
Figure~\ref{fig:exp28-rate-of-change} shows how each latent variable affects the decoder coefficients 1 and 2 in the prediction, which are associated with POD modes 1 and 2 in the prediction.
For both data modes, the largest contribution comes from the latent variable 13, followed by the latent variable 28.
The numbering changes with the seeds and gradient update algorithms, but the conclusions do not. 
For the output of the AE to represent mainly the data mode 1 and 2, we set all latent variables (except for $\vec{Z}_{13,:}$ and $\vec{Z}_{28,:}$) to $\vec{0}$. This yields the filtered latent space $\mat{Z}_f$ such that 
\begin{equation}
    \hat{\mat{P}}_f = F_{de}( \mat{Z}_f ),
\end{equation}
which contains mostly data modes~1 and~2. 
The results of decoding the filtered latent variables are shown in Figure~\ref{fig:exp28-filtered-rms}.
The RMS of the filtered prediction shows strong fluctuations only in the outer region. 
The PSD of the filtered prediction (Figure~\ref{fig:exp28-filtered-psd}) shows that  $\hat{\mat{P}}_f$ has a large amount of energy at $St\approx0.2$ and $St\approx0.002$. 
The frequency $St\approx0.06$ does not appear in the PSD of the filtered prediction, meaning that by keeping only the two most contributing latent variables for data modes 1 and 2, the AE no longer models the pulsation of the vortex core. 
This means that the filtering has been successful. 

\begin{figure}
    \centering
    \begin{subfigure}{0.47\textwidth}
        \includegraphics[width=\textwidth]{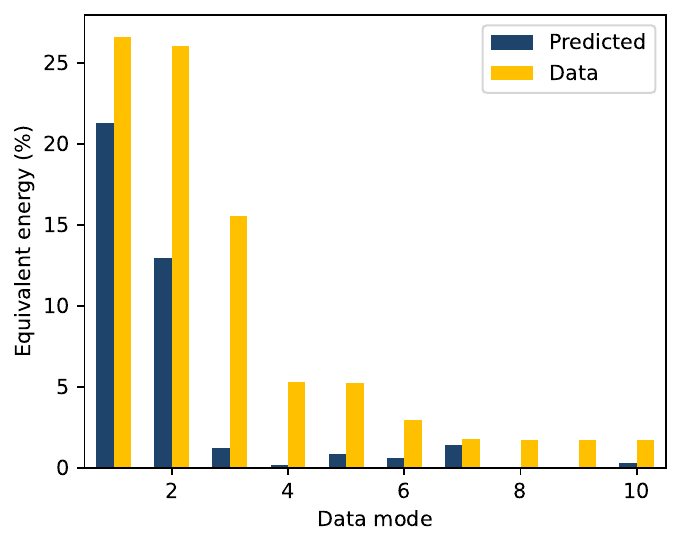}
        \caption{}\label{fig:exp28-filtered-equivalent-energy}
    \end{subfigure}
    \begin{subfigure}{0.47\textwidth}
        \includegraphics[width=\textwidth]{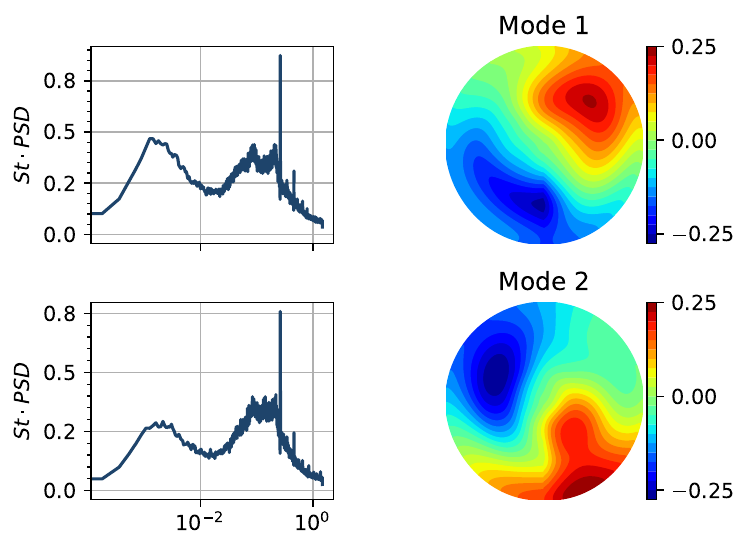}
        \caption{}\label{fig:exp28-filtered-pod-modes}
    \end{subfigure}
    \caption{
       POD on the prediction obtained with the AE with 28 latent variables from the filtered latent variables.
        a) Equivalent energy of the filtered prediction $\hat{\mat{P}}_f$ compared to the POD eigenvalues of the reference data. Amongst the five most energetic POD modes, the filtered prediction shows a large amount of energy only in data modes~1 and~2, which is the goal we set.
        b) POD modes 1 and 2 of $\hat{\mat{P}}_f$, which contain over 99\% of the flow energy of $\hat{\mat{P}}_f$. These two modes show the same frequency peak at $St\approx0.2$ and have anti-symmetric spatial structures. These two modes represent vortex shedding. 
    }\label{fig:exp28-filtered-overall}
\end{figure}
The equivalent energy in Figure~\ref{fig:exp28-filtered-equivalent-energy} further shows that the filtered prediction represents only the data modes 1 and 2, which is the objective that we set.
Figure~\ref{fig:exp28-filtered-pod-modes} shows the two leading POD modes of the filtered prediction.
The modes are approximately antisymmetric with frequency peaks at $St\approx0.2$ and $St\approx0.002$, which represent vortex shedding.
The structures in Figure~\ref{fig:exp28-filtered-pod-modes} are not perfectly antisymmetric due to information being lost by removing 26 out of 28 latent variables.
The two leading modes contain over $99\%$ of the energy in the filtered prediction, showing that the filtering has significantly compressed the spatial information.
Combining the energy, the structures and frequency content, we conclude that the only coherent structures in the filtered prediction are modes associated with vortex shedding.
By selecting latent variables based on contribution to the decoder coefficients, we have successfully filtered the coherent structures of the output of the AE.\@
Thus, we have shown decoder decomposition as a viable method for selecting latent variables based on coherent structures of the flow for an AE with a large latent dimension. 
This is useful for filtering spurious or unwanted latent variables, as well as singling out specific modes of interest.

\section{Conclusions}\label{sec:conclusion}

We propose the decoder decomposition to help practitioners design and interpret the latent space of nonlinear autoencoders.
The decoder decomposition is a post-processing method, which ranks the latent variables of an autoencoder. 
Two types of autoencoders are considered: the standard autoencoders (AE) with one decoder and one encoder, and the mode-decomposing autoencoders (MD-AEs) with one decoder per latent variable.
First, we apply the decoder decomposition to analyse the latent space of the two-dimensional unsteady laminar cylinder wake.
Both AEs and MD-AEs are applied to the laminar wake and their results are compared.
By analysing different dimensions of the latent space, we show that the dimension of the latent space significantly impacts the physical meanings of the latent variables even if the reconstruction errors remains unaffected.
Second, we apply the decoder decomposition to AEs trained with  wind-tunnel experimental data of a three-dimensional turbulent wake past a bluff body. 
We show that increasing the size of the decoder  has a similar effect to increasing the number of latent variables when the number of latent variables is small.
Third, we apply the decoder decomposition to rank and select the latent variables that are associated with the  vortex shedding structures. 
We apply the average rate of change to rank the latent variables based on their contribution to the data modes~1 and~2, which correspond to vortex shedding.
Finally, we filter the latent space to eliminate the unwanted modes. The output  contains only the two coherent structures associated with vortex shedding, which verifies the method. 
The decoder decomposition is a simple, yet robust, method for the ranking and selection of latent variables based on the flow physics and coherent structures of interest.

\FloatBarrier
\begin{Backmatter}
\paragraph{Acknowledgments}
We thank G. Rigas and J. Morrison for providing the experimental data in section~\ref{sec:data-exp}.
\paragraph{Funding Statement}
We acknowledge funding from the Engineering and Physical Sciences Research Council, UK and financial support from the ERC Starting Grant PhyCo 949388. L.M. is also grateful for the support from the grant EU-PNRR YoungResearcher TWIN ERC-PI\_0000005.
\paragraph{Competing Interests}
None.
\paragraph{Data Availability Statement}
Codes and data are accessible via GitHub: \url{https://github.com/MagriLab/MD-CNN-AE}.
\paragraph{Ethical Standards}
The research meets all ethical guidelines, including adherence to the legal requirements of the study country.
\paragraph{Author Contributions}
Y. Mo collected the numerical data, performed the analysis and wrote the paper. T. Traverso helped with the analysis and editing of the paper. L. Magri conceived the objectives and methods, and wrote and edited the paper.

\bibliographystyle{apalike}
\bibliography{reference.bib}
\end{Backmatter}
\clearpage

\begin{appendix}
\section*{Appendices}
\section{Autoencoder layers}\label{sec:appendix-layers}

The convolutional autoencoders are trained with the laminar wake dataset.
All convolutional AEs and MD-AEs are built with encoders described in Table~\ref{tab:appendix-cnn-encoder} and decoders described in  Table~\ref{tab:appendix-cnn-decoder}. 
The encoders are identical except for the latent dimension $N_z$.
For all convolutional AEs, the decoders are identical except for $N_z$.
For all convolutional MD-AEs, the input to the decoder has shape~$(1)$,  specified in Table~\ref{tab:appendix-cnn-decoder}.
When training on the wind-tunnel dataset, AEs with feedforward layers are used (tables~\ref{tab:appendix-mlp-encoder} and~\ref{tab:appendix-mlp-decoder}).
The large decoder-only network (Table~\ref{tab:appendix-large-decoder}) is only used in section~\ref{sec:results-exp-small-nz}.
The hyperparameters used by all networks  can be found in Table~\ref{tab:appendix-hyperparameters}.
\begin{table}[tbh]
    \centering
    \TBL{
    \caption{
        The encoder used in all convolutional AEs and MD-AEs.
    }
    }{
        \begin{tabular}{|c|c|c|}
            \hline
            \textbf{Layers} & \textbf{Output shape} & \textbf{Notes}\\
            \hline
            input & (200,129,1) & \\
            conv2d & (200,129,8) & 3-by-3 filter \\
            maxpooling & (100,65,8) & \\
            conv2d & (100,65,16) & 3-by-3 filter \\
            maxpooling & (50,33,16) & \\
            conv2d & (50,33,32) & 3-by-3 filter \\
            maxpooling & (25,17,32) & \\
            flatten & (13600) & \\
            dense & ($N_z$) & output shape is (1) if MD-AE\\
            \hline
        \end{tabular}
    }\label{tab:appendix-cnn-encoder}
\end{table}

\begin{table}[tbh]
    \centering
    \TBL{
    \caption{
        The decoder used in all convolutional AEs and MD-AEs.
    }
    }{
        \begin{tabular}{|c|c|c|}
            \hline
            \textbf{Layers} & \textbf{Output shape} & \textbf{Notes}\\
            \hline
            input & ($N_z$) & shape is (1) if MD-AE \\
            dense & (13600) & \\
            reshape & (25,17,32) & \\
            upsampling & (50,33,32) & bi-linear interpolation \\
            conv2d & (50,33,16) & 3-by-3 filter \\
            upsampling & (100,65,16) & bi-linear interpolation \\
            conv2d & (100,65,8) & 3-by-3 filter \\
            upsampling & (200,129,8) & bi-linear interpolation \\
            conv2d & (200,129,1) & 3-by-3 filter \\
            \hline
        \end{tabular}
    }\label{tab:appendix-cnn-decoder}
\end{table}

\begin{table}[tbh]
    \centering
    \TBL{
    \caption{
        The encoder used in feedforward AEs.
    }
    }{
        \begin{tabular}{|c|c|c|}
            \hline
            \textbf{Layers} & \textbf{Output shape} & \textbf{Notes}\\
            \hline
            input & (64) & \\
            dense & (128) & \\
            batch\_normalisation & (128) & \\
            dropout & (128) & \\
            dense & (256) & \\
            batch\_normalisation & (256) & \\
            dropout & (256) & \\
            dense & (256) & \\
            batch\_normalisation & (256) & \\
            dropout & (256) & \\
            dense & (128) & \\
            batch\_normalisation & (128) & \\
            dropout & (128) & \\
            dense & (64) & \\
            batch\_normalisation & (64) & \\
            dropout & (64) & \\
            dense & ($N_z$) & \\
            \hline
        \end{tabular}
    }\label{tab:appendix-mlp-encoder}
\end{table}

\begin{table}[tbh]
    \centering
    \TBL{
    \caption{
        The decoder used in feedforward AEs.
    }
    }{
        \begin{tabular}{|c|c|c|}
            \hline
            \textbf{Layers} & \textbf{Output shape} & \textbf{Notes}\\
            \hline
            input & ($N_z$) & \\
            dense & (64) & \\
            batch\_normalisation & (64) &\\
            dropout & (64) & \\
            dense & (128) & \\
            batch\_normalisation & (128) &\\
            dropout & (128) & \\
            dense & (256) & \\
            batch\_normalisation & (256) &\\
            dropout & (256) & \\
            dense & (256) & \\
            batch\_normalisation & (256) &\\
            dropout & (256) & \\
            dense & (128) & \\
            batch\_normalisation & (128) &\\
            dropout & (128) & \\
            dense & (64) & \\
            \hline
        \end{tabular}
    }\label{tab:appendix-mlp-decoder}
\end{table}

\begin{table}
    \centering
    \TBL{
    \caption{
        The hyperparameters used in all networks. CosineDecayRestarts is a built-in tensorflow learning rate schedule \citep{tensorflow2015-whitepaper,loshchilov2017SGDRStochasticGradient}.
    }
    }{
        \begin{tabular}{ | p{3.5cm} | p{2.1cm} | p{3cm} | p{2.9cm} |} 
        \hline
            \textbf{Which network}\newline(table/figure number(s) of the network(s) using these hyperparameters)
                & MD-AEs and AEs for the unsteady laminar wake \newline(Table~\ref{tab:appendix-cnn-encoder},~\ref{tab:appendix-cnn-decoder})
                & AEs used to search for suitable latent dimension\newline(Figure~\ref{fig:exp-convergence})
                & AEs for the wind-tunnel turbulent wake\newline(Table~\ref{tab:appendix-mlp-encoder},~\ref{tab:appendix-mlp-decoder} and~\ref{tab:appendix-large-decoder})
                \\
        \hline
            \textbf{Learning rate} & $0.001$ & $0.004$ & $0.0022$\\
            \textbf{Learning rate schedule} & n/a & CosineDecayRestarts  & CosineDecayRestarts\\
            \textbf{Activation function} & $\tanh$ & $\tanh$ & $\tanh$ \\
            \textbf{Regularisation} & $0.0$ & $0.0$ & $0.00003$ \\
            \textbf{Dropout rate} & $0.0\%$ & $0.0\%$ & $1.4\%$\\
        \hline
        \end{tabular}
    }\label{tab:appendix-hyperparameters}
\end{table}

\begin{table}[tbh]
    \centering
    \TBL{
    \caption{
        Large decoder used in section~\ref{sec:results-exp-small-nz}
    }
    }{
        \begin{tabular}{|c|c|c|}
            \hline
            \textbf{Layers} & \textbf{Output shape} & \textbf{Notes}\\
            \hline
            input & (2) & The large decoder is only used with $N_z=2$\\
            dense & (64) & \\
            batch\_normalisation & (64) &\\
            dropout & (64) & \\
            dense & (256) & \\
            batch\_normalisation & (256) &\\
            dropout & (256) & \\
            dense & (256) & \\
            batch\_normalisation & (256) &\\
            dropout & (256) & \\
            dense & (256) & \\
            batch\_normalisation & (256) &\\
            dropout & (256) & \\
            dense & (256) & \\
            batch\_normalisation & (256) &\\
            dropout & (256) & \\
            dense & (64) & \\
            \hline
        \end{tabular}
    }\label{tab:appendix-large-decoder}
\end{table}
\FloatBarrier

\section{POD and linear AE}\label{sec:appendix-linear}

Performing POD is equivalent to solving the quadratic optimization problem \citep{fukami2021AEandROM}
\begin{equation}
    \mat{\Phi} = \arg \min_{\mat{\Phi}^*} ||\mat{Q}-\mat{\Phi}^* \mat{\Phi}^{*T} \mat{Q}||_2^2 .
    \label{eq:method-linear-podoptimisation}
\end{equation}
Let us consider a linear AE in which the number of latent variables is equal to the number of grid points in the input, i.e. $N_z = N$. 
A linear AE is an AE with linear activation functions.
If $N_z < N$, the linear AE recovers only the $N_z$ most energetic POD modes in an $L_2$-norm sense.
For a linear AE without biases, we express the training as 
\begin{equation}
    \boldsymbol{\omega}^* = \arg\min_{\boldsymbol{\omega}} \| \mat{Q} - \hat{\mat{W}}\mat{W}\mat{Q} \|_2^2,
    \label{eq:method-linear_standard_nobias_optimise}
\end{equation}
which has the solution
\begin{equation}
    \hat{\mat{W}} \mat{W} = \mat{I},
    \label{eq:method-linear-inverse_W}
\end{equation}
where $\mat{W}$ and $\hat{\mat{W}}$ are the weights of the encoder and the decoder of the linear AE. 
These matrices are not necessarily orthogonal.
To obtain the POD solution, we seek orthogonal $\hat{\pmb{W}}$ and $\pmb{W}$.
\citet{baldi1989MLP_PCA} showed that for a linear AE without biases, the only solution is the POD solution, whereas all the remaining critical points are saddle points.
However, there are saddle points in the optimisation of~\eqref{eq:method-linear_standard_nobias_optimise}.
We apply $L_2$-regularization to the weights for the linear AE to converge to the POD solution, which is equivalent to minimizing the Frobenius norm of weights, denoted  $\|\mat{W}\|_F$.
When the autoencoder is linear, applying $L_2$-regularization of weights is known as applying a ``soft constraint of orthogonality'' \citep{bansal2018CanWeGain,xie2017AllYouNeed}.
Training the linear AE with $L_2$-regularization is the solution of  
\begin{equation}
    \boldsymbol{\omega}^* = \arg\min_{\boldsymbol{\omega}} || \mat{Q}- \hat{\mat{Q}} ||_2^2 + \gamma \left( \|\mat{W}\|_F^2 + \|\hat{\mat{W}}\|_F^2 \right ),
    \label{eq:method-linear-frobenius-loss}
\end{equation}
where $\gamma \geq 0$ is the regularization factor.
The only solution to the optimization problem~\eqref{eq:method-linear_standard_nobias_optimise} which minimises~\eqref{eq:method-linear-frobenius-loss} is
\begin{equation}
    \mat{W} \mat{W}^T = \mat{I}
\end{equation}
where $\mat{I}$ is the identity matrix. Therefore, $
    \hat{\mat{W}} = \mat{W}^T$, 
i.e., the decoder and encoder matrices are orthogonal. 
A similar analysis can be applied to a linear MD-AE to show that each decomposed field is a POD mode.

\section{Physical interpretation is more difficult when unsuitable latent dimension is used}\label{sec:result-cylinder-ae-wrongnz}

\begin{figure}
    \centering
    \begin{subfigure}{0.4\textwidth}
        \includegraphics[width=\textwidth]{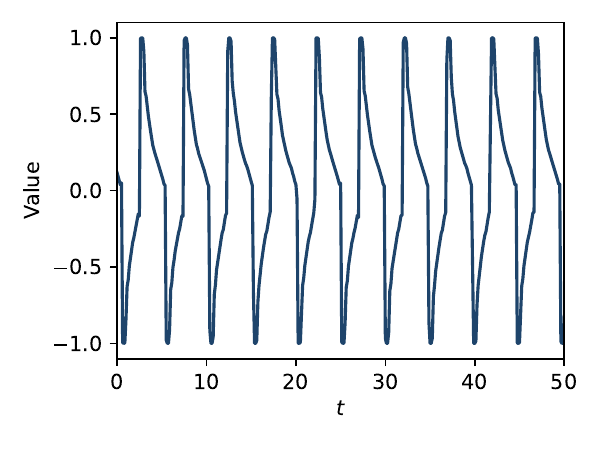}
        \caption{}\label{fig:cylinder-ae1-timetrace}
    \end{subfigure}
    \begin{subfigure}{0.4\textwidth}
        \includegraphics[width=\textwidth]{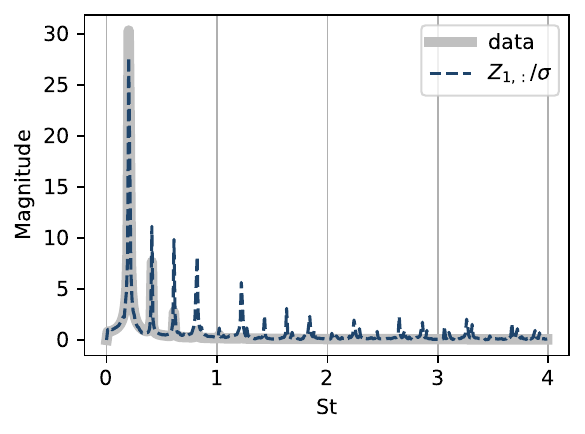}
        \caption{}\label{fig:cylinder-ae1-freq}
    \end{subfigure}
    \caption{
        Results from the AE trained with the laminar wake dataset with 1 latent variable.
        a) The latent variable approximates the discontinuity caused by the angle moving from $2\pi$ to $0$. 
        b) The frequency spectrum of the latent variable, normalized by its standard deviation, compared with the data frequencies. The latent variables contain frequencies that do not exist in the dataset.
    }\label{fig:cylinder-ae1}
\end{figure}

\begin{figure}
    \centering
    \begin{subfigure}{0.36\textwidth}
        \includegraphics[width=\textwidth]{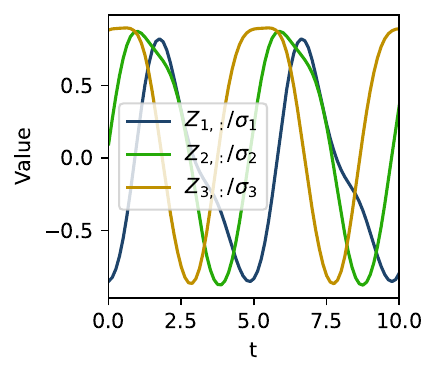}
        \caption{}\label{fig:cylinder-ae3-timetrace}
    \end{subfigure}
    \begin{subfigure}{0.32\textwidth}
        \includegraphics[width=\textwidth]{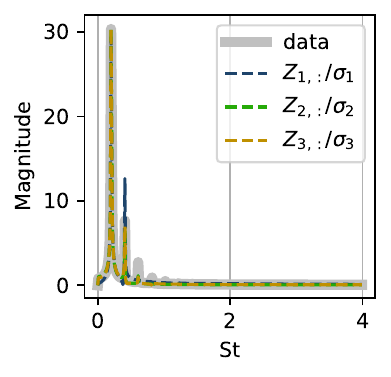}
        \caption{}\label{fig:cylinder-ae3-freq}
    \end{subfigure}
    \begin{subfigure}{0.29\textwidth}
        \includegraphics[width=\textwidth]{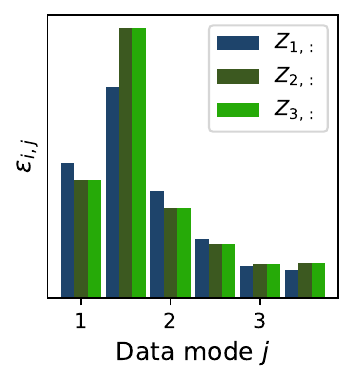}
        \caption{}\label{fig:cylinder-ae3-rateofchange}
    \end{subfigure}
    \caption{
        Results from the AE trained with the laminar wake dataset with three latent variables.
        a) The time trace of the latent variables.
        b) The frequency spectrum of the latent variables, normalized by their standard deviations. The latent variables contain the vortex shedding and the first harmonic frequency.
        c) The averaged rate of change of decoder coefficients due to latent variables of an AE with three latent variables. 
        Latent variables $\vec{Z}_{2,:}$ and $\vec{Z}_{3,:}$ have the same contributions towards the first six data modes, meaning that both latent variables carry the same information.
    }\label{fig:cylinder-ae3-results}
\end{figure}

In section~\ref{sec:result-cylinder-ae}, we show that the MSE of the AEs converges for $N_z=2$, and that at least two latent variables are needed to describe the periodic behaviour.
However, it is difficult to know the theoretical number of latent variables to use in common machine learning tasks. The common approach in machine learning is to treat the latent dimension as a hyperparameter of the networks.
By using the laminar wake as an example, we  discuss how the choice of the latent variables affects the physical interpretation of the latent variables.
First, we show the results from an AE with $N_z=1$ in Figure~\ref{fig:cylinder-ae1}.
The latent variable is periodic in time, but the changes are not smooth (Figure~\ref{fig:cylinder-ae1-timetrace}) and have frequency peaks at high frequencies not present in the data (Figure~\ref{fig:cylinder-ae1-freq}).
When a single variable (the angle) is used to represent the periodic behaviour of the dataset $\mat{U}$, there is a discontinuity when the angle moves from $2\pi$ to $0$.
The latent variable is forced to use higher frequencies to try to approximate the discontinuity.
These higher frequencies are the results of numerical approximation and are unphysical.
In an AE with $N_z=3$, the latent variables contain the first harmonic in addition to the vortex shedding frequency (Figure~\ref{fig:cylinder-ae3-freq}), which is present in the data but not in the first two data time coefficients (Figure~\ref{fig:cylinder-data-pod-results}, right panel) or the latent variables from an AE with two latent variables (Figure~\ref{fig:cylinder-ae2-latentbasicb}).
We discussed (section~\ref{sec:result-cylinder-ae}) that the additional frequency peak is not necessary for the accurate reconstruction of the dataset.

The average  rate of change shows that latent variables $\vec{Z}_{2,:}$ and $\vec{Z}_{3,:}$ contribute equally to the first six data modes (Figure~\ref{fig:cylinder-ae3-rateofchange}).
Equal contribution to all data modes indicates that the two latent variables have the same role in representing the data, which means that the information carried by one of the latent variables is  duplicate information contained in the other latent variables.  The three latent variables are not suitable because they lead to redundant information.
All three latent variables contribute similarly to the first six data modes, making the selection of latent variables more difficult.
In the present case, the three latent variables form a group to represent the counter-rotating vortices, even though two latent variables should be, in principle, sufficient.

Both the AE with $N_z=3$ and the MD-AE with $N_z=2$ (section~\ref{sec:result-cylinder-mdae}) have latent variables with an additional frequency peak compared to the latent variables of the AE with $N_z=2$ (section~\ref{sec:result-cylinder-ae}), even though there is no substantial difference in the reconstruction error.
Additionally, unphysical frequencies also arise from numerical approximations when using AE with $N_z=1$.
Therefore, a design consideration of AEs for nonlinear reduced-order modelling is to use a latent dimension that results in the least number of frequency peaks in the latent variables whilst maintaining the reconstruction accuracy of the AEs.

\makeatletter
\def\thenomenclature{%
  \@ifundefined{chapter}%
  {
    \section{\nomname}\label{nom}
    \if@intoc\addcontentsline{toc}{subsection}{\nomname}\fi%
  }%
  {
    \chapter{\nomname}
    \if@intoc\addcontentsline{toc}{chapter}{\nomname}\fi%
  }%

  \nompreamble%
  \list{}{%
    \labelwidth\nom@tempdim
    \leftmargin\labelwidth
    \advance\leftmargin\labelsep
    \itemsep\nomitemsep
    \let\makelabel\nomlabel}}
\makeatother

\makenomenclature%

\nomenclature[M]{$\mat{Q}$}{Snapshot matrix}
\nomenclature[M]{$\mat{\Lambda}$}{Diagonal matrix of eigenvalues, sorted from largest to smallest}
\nomenclature[M]{$\mat{\Phi}$}{POD modes}
\nomenclature[M]{$\mat{C}$}{Correlation matrix}
\nomenclature[M]{$\mat{W}_P$}{Weight matrix for the snapshot matrix for POD}
\nomenclature[M]{$\mat{A}$}{Time coefficients for POD modes}
\nomenclature[M]{$\mat{I}$}{Identity matrix}
\nomenclature[M]{$\tilde{\mat{Q}}$}{Reconstructed snapshot matrix $\mat{Q}$ using $N_m$ POD modes.}
\nomenclature[M]{$\mat{U}$}{The laminar wake dataset, consists of the streamwise velocity of the simulated laminar wake on a two-dimensional grid.}
\nomenclature[M]{$\mat{P}$}{The wind-tunnel wake dataset, consists of the pressure measurements on a two-dimensional polar grid.}
\nomenclature[M]{$\mat{Y}$}{An example data matrix used for defining the autoencoders and related methods.}
\nomenclature[M]{$\mat{Z}$}{Matrix of latent variables}
\nomenclature[M]{$\mat{B}$}{Matrix of decoder coefficients}
\nomenclature[M]{$\mat{M}^i$}{The $i$-th decomposed field of an MD-AE.}
\nomenclature[M]{$\hat{\mat{\Lambda}}$}{The equivalent energy of the output of an AE.}
\nomenclature[M]{$\hat{\mat{\Lambda}}^i$}{The equivalent energy of the $i$-th decomposed field of a MD-AE.}

\nomenclature[S]{$N$}{The number of measured variables in the dataset}
\nomenclature[S]{$N_t$}{Number of time steps}
\nomenclature[S]{$N_m$}{Number of truncated POD modes to keep for reconstruction}
\nomenclature[S]{$\vec{u}$}{Velocities}
\nomenclature[S]{$p$}{Pressure}
\nomenclature[S]{$N_u$}{Number of velocity components}
\nomenclature[S]{L}{Domain length}
\nomenclature[S]{$U_\infty$}{Inlet velocity}
\nomenclature[S]{D}{Diameter of the cylinder/axisymmetric body}
\nomenclature[S]{f}{Frequency}
\nomenclature[S]{T}{Period of vortex shedding}
\nomenclature[S]{$F_{en}$}{Function composition that represents the encoder}
\nomenclature[S]{$F_{de}$}{Function composition that represents the decoder}
\nomenclature[S]{$\vec{\omega}$}{parameters of a network}
\nomenclature[S]{$\hat{*}$}{Dataset * reconstructed by an autoencoder}
\nomenclature[S]{$\sigma$}{Standard deviation}

\nomenclature[A,2]{POD}{Proper Orthogonal Decomposition}
\nomenclature[A,3]{$\sym{lam}$}{Laminar}
\nomenclature[A,3]{$\sym{exp}$}{Experimental}
\nomenclature[A,2]{PSD}{Power Spectral  Density}
\nomenclature[A,1]{AE}{A standard autoencoder}
\nomenclature[A,1]{MD-AE}{A mode-decomposing autoencoder}
\nomenclature[A,2]{MSE}{Mean Square Error}

\nomenclature[N]{$\sym{Re}$}{Reynolds number}
\nomenclature[N]{$\sym{St}$}{Strouhal number}

\printnomenclature

\end{appendix}
\end{document}